%% file: main.tex
\documentclass[letterpaper]{article}
\usepackage{uai2019}
\usepackage[margin=1in]{geometry}

\usepackage{times}

\usepackage[english]{babel}
\usepackage{blindtext}
\title{Towards Robust Relational Causal Discovery}

\author{Sanghack Lee\\
Causal AI Laboratory\\
Department of Computer Science\\
Purdue University\\
West Lafayette, IN 47907\\
\texttt{lee2995@purdue.edu}
\And
Vasant Honavar\\
Artificial Intelligence Research Laboratory\\
College of Information Sciences and Technology\\
Pennsylvania State University\\
University Park, PA 16802\\
\texttt{vhonavar@psu.edu}
}

\usepackage{times}
\usepackage{url}
\usepackage[utf8]{inputenc}
\usepackage{graphicx}
\usepackage{amsmath}
\usepackage{booktabs}

\usepackage{natbib}
\usepackage{bm}
\usepackage{lipsum}
\usepackage{eucal}
\usepackage{multirow}
\usepackage{float}
\usepackage{enumitem}
\usepackage{algorithm}
\usepackage[noend]{algpseudocode}
\usepackage{xcolor}
\usepackage{array}
\usepackage{mathrsfs}
\usepackage{amsmath}
\usepackage{amsthm}
\usepackage{amssymb}
\usepackage[T1]{fontenc}    
\usepackage{booktabs}       
\usepackage{amsfonts}       
\usepackage{nicefrac}       
\usepackage{microtype}      
\usepackage{pifont}
\usepackage{subcaption}
\usepackage{cleveref}

\usepackage{stmaryrd}
\usepackage{tikz}

\usepackage{thmtools}
\usepackage{thm-restate}
\usepackage[title]{appendix}
\usepackage{soul}

\usepackage{sansmath}
\usepackage{tikz}
\usetikzlibrary{arrows.meta, positioning,calc,fit,shapes}
\tikzset{%
>={Latex[width=1.1mm,length=1.6mm]},
        base/.style = {rectangle, draw=black, rounded corners, 
                       minimum width=0.4cm, minimum height=0.3cm,
                       text centered, font=\sffamily},
elabel/.style = {midway,right ,opacity=0,text opacity=1},
attr/.style = {rectangle, draw=black, rounded corners},
entc/.style = {rectangle, draw=black},
skit/.style={minimum width=8mm, minimum height=4mm, inner sep=0.5mm, draw,align=left,text width=10mm},
relc/.style = {diamond, draw=black,minimum height=4mm,minimum width=9mm},
relcs/.style = {diamond, draw=black,minimum height=0.3mm,minimum width=0.3mm},
attrs/.style={rectangle,rounded corners=0.6mm, draw,inner sep=0.2mm,minimum width=2.2mm,minimum height=2.7mm}
}

\definecolor{betterred}{RGB}{228,26,28}
\definecolor{betterblue}{RGB}{55,126,184}
\definecolor{bettergreen}{RGB}{77,175,74}
\definecolor{betterpurple}{RGB}{152,78,163}

\theoremstyle{definition}
\newtheorem{defn}{Definition}

\Crefname{equation}{Eq.}{Eqs.}
\Crefname{figure}{Fig.}{Figs.}
\Crefname{section}{Sec.}{Secs.}
\Crefname{prop}{Prop.}{Props.}
\Crefname{lem}{Lemma}{Lemmas.}
\Crefname{thm}{Thm.}{Thms.}
\Crefname{algorithm}{Alg.}{Algs.}
\Crefname{table}{Tab.}{Tabs.}

\newcommand\Perp{\protect\mathpalette{\protect\independenT}{\perp}} \def\independenT#1#2{\mathrel{\rlap{$#1#2$}\mkern3mu{#1#2}}}
\newcommand\notperp{\mathbin{\not\Perp}}

\crefname{appsec}{Appendix}{Appendices}

\newcommand\schema{\mathcal{S}}
\newcommand\GG{\mathcal{G}}
\newcommand\MM{\mathcal{M}}
\newcommand\RR{\bm{R}}
\newcommand\EEE{\bm{E}}
\newcommand\DD{\bm{D}}
\newcommand\AAA{\bm{A}}

\newcommand\card{\mathsf{card}}
\newcommand{\VV}[1] {\mathcal{V}_{#1}}

\newcommand{\VVV}{\bm{V}}
\newcommand{\WW}{\bm{W}}
\newcommand{\XX}{\bm{X}}

\newcommand{\vvv}{\bm{v}}

\newcommand{\xx}{\bm{x}}
\newcommand{\yy}{\bm{y}}

\newcommand\GGG{{\mathcal{G}_{\sigma}^{\mathcal{M}}}}

\newcommand{\braces}[2][]{#1\{#2 #1\}}

\newcommand{\brackets}[2][]{#1[#2 #1]}
\newcommand{\parens}[2][]{#1(#2 #1)}

\newcommand{\verts}[2][]{#1\lvert#2 #1\rvert}
\newcommand{\angles}[2][]{#1\langle#2 #1\rangle}

\newcommand{\set}[2][]{\braces[#1]{#2}}

\newcommand{\tuple}[2][]{\angles[#1]{#2}}

\algrenewcommand\algorithmicindent{0.7em}

\definecolor{cBa}{RGB}{239,243,255}
\definecolor{cBb}{RGB}{189,215,231}
\definecolor{cBc}{RGB}{107,174,214}
\definecolor{cBd}{RGB}{49,130,189}
\definecolor{cBe}{RGB}{8,81,156}

\begin{document}

\maketitle

\input{abstract}

\section{INTRODUCTION}
\label{sec:introduction}
\input{introduction}

\section{PRELIMINARIES}
\label{sec:preliminaries}
\input{preliminaries}

\section{RELATIONAL CAUSAL DISCOVERY}
\label{sec:rpcd}
\input{rpcd}

\section{TESTING RCI USING A CI TEST}
\label{sec:cit}
\input{cit}

\section{ROBUST RELATIONAL CAUSAL DISCOVERY}
\label{sec:rcd}
\input{rcd}

\section{EXPERIMENTS AND RESULTS}
\label{sec:experiments}
\input{experiments}

\section{SUMMARY AND DISCUSSION}
\label{sec:conclusions}
\input{conclusions}
\bibliography{aaai-rrcd}
\bibliographystyle{named}

\clearpage\newpage
\section*{Appendix}
\input{appendix}

\end{document}

%% file: abstract.tex
\begin{abstract}
We consider the problem of learning causal relationships from relational data. Existing approaches rely on queries to a relational conditional independence (RCI) oracle to establish and orient causal relations in such a setting. In practice, queries to a RCI oracle have to be replaced by reliable tests for RCI against available data. Relational data present several unique challenges in testing for RCI. We study the conditions under which traditional iid-based CI tests yield reliable answers to RCI queries against relational data. We show how to conduct CI tests against relational data to robustly recover the underlying relational causal structure. Results of our experiments demonstrate the effectiveness of our proposed approach.
\end{abstract}

%% file: introduction.tex

Determining causal effects from observations and experiments is a central concern of all sciences, and increasingly, of artificial intelligence, data sciences, and statistics \citep{Pearl2000book,Spirtes2000book,Rubin1974a}. 
Causal inference allows one to elicit causal effects among variables given partial knowledge or assumptions about the data generating process within a domain of interest, often represented by a {\em causal graph}, a directed acyclic graph where the nodes represent variables of interest and directed edges denote direct causes. Causal discovery is concerned with obtaining causal knowledge by analyzing data obtained from the system of interest (which can be a model, population, or nature). However, because data provides, at best, only partial information about the underlying system, causal assumptions about the world are essential for causal discovery. Consequently, different algorithms for causal discovery often embody different assumptions about the underlying world \citep{verma:pea90a,spirtes:etal95}.

Most existing causal discovery algorithms are designed to learn a causal graph over the variables $\VVV$ where data consists of independent and identically distributed (iid) instances, where each instance corresponds to an instantiation $\vvv$ of the variables. Conditional independence relations between variables implicit in the data, a sample of the joint distribution $P(\vvv)$, can partially reveal the underlying causal graph. However, data in many real-world settings violate the iid assumption because they are generated by a system of \emph{interacting} objects e.g., a collaboration network, social network, or entities connected by {\em relations} stored in relational databases. Hence, there is growing interest in methods for learning causal models from relational data. 
\citet{maier2010rpc} considered a causal model for relational domains, and devised an algorithm called RPC (Relational PC);
\citet{maier2013rcd} introduced the Relational Causal Model (RCM), a revised version of their previous model, and proposed the Relational Causal Discovery algorithm, for learning a RCM from data\footnote{Depending on context, we will use RCD, which stands for {\em relational causal discovery}, to refer to the problem, or the specific solution proposed by \citet{maier2013rcd}};
\citet{lee2016rcdl} introduced RCD-Light, a more efficient version of the RCD algorithm; The same authors 
\citeyearpar{Lee2016UAI} proposed RpCD, a relational CI oracle based RCD algorithm, that is sound, and unlike RCD and RCD-Light, also {\em complete}. Unfortunately, this body of work largely falls short of offering a practical solution to RCD. One main reason has to do with the fact that, in practice, the relational CI (RCI) oracle must be replaced by reliable RCI tests; however, most of the existing CI tests do not account for the relational structure underlying relational data, and hence fail to produce reliable answers for RCI queries. Although several CI tests for some types of non-iid data have been proposed in the literature, e.g., the test proposed by \citet{Flaxman2016tist}, which has been shown to work well for temporal, spatial, or undirected graph-structured data, such tests are not directly applicable to relational data. \citet{LH2017KRCIT} proposed KRCIT, a suite of graph kernel based relational CI tests, which can reduce the false positive answers to RCI queries resulting from the violation of the iid assumption when a CI test that designed for iid data is naively applied to relational data, they suffer from low power which can result in failure to detect relational CI.

\noindent
{\bf Contributions} We propose a relational causal discovery algorithm that effectively works with the available (necessarily imperfect) relational CI tests.
Specifically:
1) 
We identify the conditions under which CI tests that assume iid data can reliably answer relational CI queries, and show how the resulting insights can be exploited by algorithms for RCD;
2) We examine the consequences of replacing a relational CI oracle by relational CI tests from the perspective of relational causal discovery, and propose ways to increase the robustness RCD algorithms that use imperfect relational CI tests.

%% file: preliminaries.tex

\input{tikz_prelim}

The Relational Causal Model (RCM) \citep{maier2013rcd} marries a relational schema \citep{chen1976entity} used by relational databases representing the relational structure of the domain with the causal Bayesian network (CBN) \citep{Pearl2000book} used to represent the structure and parameters of causal models.
We borrow many of the notations introduced in the existing literature on RCMs \citep{maier2013rcd,Lee2016UAI}. 
We use an uppercase letter, e.g., $X$, to denote a variable, and the corresponding lower case letter, e.g., $x$, to denote its realization. We use
bold letters, e.g., $\XX$ or $\xx$, to represent sets, and 
calligraphic letters to represent complex mathematical objects.
We use the kinship notation, $pa$, $ch$, $an$, $de$, for graphical relationships such as parents, children, ancestors, descendants. We express the CI statement that the random variables $X$ and $Y$ are conditionally independent (CI) given $Z$, i.e. that $P(Y| X,Z)$ can be expressed
as $P(Y| Z)$, by $X\Perp Y \mid Z$.
Throughout the paper, we will make use of examples adapted from \citep{maier2014thesis}.

\noindent
{\bf Relational Domain}
A relational schema $\schema$ defines how entities interact within a given domain of interest where 
 $\schema = \tuple{\EEE,\RR,\AAA,\card}$ --- 
a set of entity classes $\EEE$, relationship classes $\RR$, attribute classes $\AAA$, and 
cardinality constraints (on the number of entities that can participate in a relationship, i.e, one, many).
See \Cref{fig:schema} for a concrete example. In this domain, there are 3 entity classes, Employee, Product, and Business Unit (unit for short), and 2 relationship classes, Develops and Funds with their attribute classes shown using rounded rectangles. We will refer to the item classes using the initial letter of their names (\textsf{E} for Employee etc.): \textsf{E}, \textsf{P}, \textsf{B}, \textsf{D}, and \textsf{F}. Small \textsf{m} near the line between \textsf{E} and \textsf{D} specifies that an employee can develop many products; and a unit can fund many products but a product can be funded by only one unit.

A relational skeleton, denoted by $\sigma\in \Sigma_{\schema}$, is one of possible realizations of the given relational schema $\schema$ where $\Sigma_{\schema}$ is a set of all possible relational skeletons (realizations) of the schema. We denote by $\sigma(B)$ a set of items in $\sigma$ corresponding to an item class $B$. Attribute value $x$ of an item $i$ is denoted by $i.x$. Entities and relationships form a bipartite graph satisfying the constraints imposed by the definition of the relationship classes and the cardinality constraints. If $E\in \EEE$ participates in $R\in\RR$ with cardinality `one', then $\verts{\set{r\mid (e,r)\in \sigma, r\in\sigma(R)}}\leq 1$ for every $e\in \sigma(E)$. Relational data is then a tuple of the network structure of the relational skeleton and the values of attributes of the items.
With the example schema, see \Cref{fig:skeleton} for a relational skeleton where there are 5 employees, 5 products, and 2 units. The illustration hides relationship items: The edge between $\mathsf{e}_1$ and $\mathsf{p}_1$ represents the existence of a relationship item $\mathsf{d}_{e_1,p_1}\in \sigma(\mathsf{D})$, which is connected to $\mathsf{e}_1$ and $\mathsf{p}_1$. The cardinality constraints impact the relational skeleton: Some employees develop multiple products and some products are developed by multiple employees; Every product is funded by at most one business unit. By definition, there is no requirement that an entity must participate in any relationship.

\noindent{\bf Relational Causal Model}
Relational Causal Model (RCM) $\MM = \tuple{\schema,\bm{D},\bm{F}}$ is defined with respect to a given relational schema $\schema$ to represent causal relations among attribute classes related via $\schema$. It consists of a set of relational dependencies $\bm{D}$, which represents causal relationships among variables (defined in the relational space, as described below), and a set of functions $\bm{F}$, which specify how attribute values are generated from their respective causes.
\Cref{fig:rcm} informally illustrates a set of 5 relational dependencies as curved edges. A relational dependency (RD) from competence to salary means that an employee's salary depends on the employee's competence. The dependency from budget to salary implies that an employee's salary (also) depends on the budget of the unit that funds a product developed by the employee. A (stochastic) function specifies how an employee's salary (say, in dollars) can be obtained from such information. 

More formally, a \emph{relational dependency} is of the form $U\to V$ where $U$ and $V$ are \textit{relational variables} such that $V$ represents the attribute class of an item class and $U$ is an attribute class defined \textit{relative} to that item class. A relational variable is of the form e.g., $P.X$. As a brief example, aforementioned dependency is expressed as $\mathsf{[EDPFB].Budget\to \brackets{E}.Salary}$. 
Such a sequence of item classes, $\mathsf{[EDPFB]}$, appearing in a relational variable is called a \textit{relational path}, which is restricted to a \emph{walk} (in a graph theoretic sense) on a relational schema from the effect's item class to the cause's item class.\footnote{This formulation is not unlike that found in early probabilistic statistical relational learning literature, e.g., \citep{Koller1999}.} A relational path defines the relationship between the attribute classes of item classes that are connected by a relational dependency.\footnote{Restricting the relationship to a path in an underlying skeleton does limit the expressivity of the resulting RCM. However such a restriction simplifies analysis of RCMs and yields a characterization of the equivalence class of a RCM, which in turn leads to a complete RCI-oracle-based RCD algorithm.} The first item class in the path is called a \emph{base} item class (or perspective), and the last item class is called a \emph{terminal} item class.
If the relational path of a relational variable is a singleton (which we call \emph{canonical}), we use the following notation, $\VV{X}=[I_X].X$ where $I_X$ is an item class owning $X$.
A relational dependency, e.g., $P.X\to \VV{Y}$, implies that the base item class of $P$ is $I_Y$ and the terminal item class is $I_X$.

A RCM $\MM$ is a specification of the causal relationships between the attributes of items of a relational skeleton of a given relational schema. Given a relational skeleton $\sigma$, the model $\MM$ is instantiated as a \emph{ground graph} $\GGG$, which is a CBN made of items' attributes. For instance, there will be a directed edge from $\mathsf{b_2.B}$ to $\mathsf{e_2.S}$ in \Cref{fig:gg} because there exists a path of items $\mathsf{[b_2,f_{p_3,b_2},p_3,d_{e_2,p_3},e_2]}$ corresponding to the relational path $\mathsf{[BFPDE]}$ where $\mathsf{f_{p_3,b_2}}$ and $\mathsf{d_{e_2,p_3}}$ are implicit in \Cref{fig:skeleton}. Formally, we use $P.X|_i^\sigma$ to denote the multi-set of $X$ of items reachable from item $i$ through a path of items in $\sigma$ corresponding to $P$. For instance,
$\mathsf{[PDE].C|_{p_3}^{\sigma}=\set{e_2.C,e_3.C,e_4.C}}.$
Then, the vertices of $\GGG$ correspond to the item attributes and edges are $\set{j.X {\to} i.Y \mid P.X{\to} \VV{Y}{\in} \DD, i{\in} \sigma(I_Y), j{\in} P|_i^\sigma}$. A ground graph plays the role of a causal model for the observed relational data. In other words, the attribute values appearing in a relational skeleton corresponds to a single instance sampled from the ground graph. 
The ground graph as shown in \Cref{fig:gg} is based on \Cref{fig:rcm,fig:skeleton}. Although there are directed edges shown upwards, $\GGG$ is acyclic with a topological order as defined by the RCM, i.e., competence, success, revenue, budget, and salary.

\noindent
{\bf Relational Conditional Independence (RCI)}
RCI (defined below) generalizes CI from the iid setting to the relational setting. Analogous to a CBN embodying a set of CI assertions, a RCM $\MM$ embodies a set of RCI assertions.
Hence, this set of RCI assertions, if available (e.g., through queries to a RCI oracle or RCI tests against data), allows us to discover the partial structure of the RCM responsible for generating the observed relational data.

Let $U$, $V$ be relational variables of a given base item class (say $B$). Let $\WW$ be a set of relational variables of the same base item class ($B$).
$U$ and $V$ are said to be RCI given $\WW$, denoted by
$U \Perp V \mid \WW$, 
if and only if
$U|_i^\sigma \Perp V|_i^\sigma \mid \WW|_i^\sigma$
for every relational skeleton $\sigma\in\Sigma_\schema$ and every item $i\in\sigma(B)$.
By the definition, RCI is a property of a RCM. That is, RCI statement considers a collection of CI statements from \emph{all} possible relational skeletons of a relational schema. However, often, only a single relational skeleton, i.e., a single instance sampled from the ground graph, is available for testing RCI. 

The following examples are intended to further illustrate the notion of RCI. The independence between a unit's budget and the unit's employees' competence given the success of the products funded by the unit can be expressed as
$\mathsf{[BFPDE].C \Perp [B].B \mid [BDP].S}$.
In contrast, consider a similar statement from a different perspective:
\begin{align}\label{eq:rciexample}
\mathsf{[EDPFB].B \not\Perp [E].C \mid [EDP].S},
\end{align}
which is because we can find a d-connection path, e.g., $\mathsf{e_1.C\to p_1.S\gets e_2.C\to p_2.S\to b_1.R\to b_1.B}$ in $\GGG$ where $\mathsf{p_1.S}$ is a collider. 
However, as mentioned earlier, it is not feasible to test whether $\mathsf{e_1.C \notperp b_1.B | p_1.S}$ from a single instance of relational data.

%% file: tikz_prelim.tex


\begin{figure*}[t]
\scriptsize
\hfill%
\begin{subfigure}{0.4\textwidth}\centering
    \begin{tikzpicture}
    \draw[-,opacity=0] (0,-1.4) -- (0,0.7);
    \node[attr] (EC) {\textsf{competence\vphantom{fg}}};
    \node[attr] (ES) [below=1mm of EC] {\textsf{salary\vphantom{fg}}};
    \node[entc,fit=(EC)(ES), draw,label=above:{\textsf{Employee}}] (E) {};

    \node[attr] (BR) [right=2.5cm of EC] {\textsf{revenue\vphantom{fg}}};
    \node[attr] (BB) [below=1mm of BR] {\textsf{budget\vphantom{fg}}};
    \node[entc,fit=(BR)(BB), draw,label=above:{\textsf{Biz-Unit}}] (B) {};
    
    \node[attr] (PS) at ($(E.east)!0.5!(B.west)$){\textsf{success\vphantom{fg}}};
    \node[entc,fit=(PS), draw,label=above:{\textsf{Product}}] (P) {};
    
    \node[relc,label=left:{\textsf{Develops}}] (D) [below left=5mm and 1mm of P] {};
    \node[relc,label=right:{\textsf{Funds}}] (F) [below right=5mm and 1mm of P] {};

    \draw[-] (D) -- (E) node[midway,label=left:{\textsf{\tiny m}}] {};
    \draw[-] (D) -- (P) node[pos=0.8,label=right:{\textsf{\tiny m}}] {};
    \draw[-] (F) -- (P);
    \draw[-] (F) -- (B) node[midway,label=right:{\textsf{\tiny m}}] {};

    \draw[opacity=0,thick,->,cBe] (EC.west) to [bend right=90,looseness=1.5] (ES.west);
    \draw[opacity=0,thick,->,cBe] (BR.10) to [bend left=60,looseness=1.5] (BB.-10);
    \end{tikzpicture}
    \caption{Relational Schema $\schema$}
    \label{fig:schema}
\end{subfigure}\hfill
\begin{subfigure}{0.5\textwidth}\centering
    \begin{tikzpicture}[node distance=3.5mm]
    \draw[-,opacity=0] (0,-1.8) -- (0,.3);
    \node[skit] (E1) {$\mathsf{e_1}$};
    \node[skit] (E2) [right=of E1] {$\mathsf{e_2}$};
    \node[skit] (E3) [right=of E2] {$\mathsf{e_3}$};
    \node[skit] (E4) [right=of E3] {$\mathsf{e_4}$};
    \node[skit] (E5) [right=of E4] {$\mathsf{e_5}$};

    \node[skit] (P1) [below=3mm of E1] {$\mathsf{p_1}$};
    \node[skit] (P2) [right=of P1]     {$\mathsf{p_2}$};
    \node[skit] (P3) [right=of P2]     {$\mathsf{p_3}$};
    \node[skit] (P4) [right=of P3]     {$\mathsf{p_4}$};
    \node[skit] (P5) [right=of P4]     {$\mathsf{p_5}$};

    \node[skit] (B1) [below=3mm of P2] {$\mathsf{b_1}$};
    \node[skit] (B2) [below=3mm of P4] {$\mathsf{b_2}$};
    
    \node[attrs,anchor=east,shift={(-5.2mm,0mm)}] (E1C) at (E1.east) {$\mathsf{c}$};
    \node[attrs,anchor=east,shift={(-5.2mm,0mm)}] (E2C) at (E2.east) {$\mathsf{c}$};
    \node[attrs,anchor=east,shift={(-5.2mm,0mm)}] (E3C) at (E3.east) {$\mathsf{c}$};
    \node[attrs,anchor=east,shift={(-5.2mm,0mm)}] (E4C) at (E4.east) {$\mathsf{c}$};
    \node[attrs,anchor=east,shift={(-5.2mm,0mm)}] (E5C) at (E5.east) {$\mathsf{c}$};

    \node[attrs,anchor=east,shift={(-.5mm,0mm)}] (E1S) at (E1.east) {$\mathsf{s}$};
    \node[attrs,anchor=east,shift={(-.5mm,0mm)}] (E2S) at (E2.east) {$\mathsf{s}$};
    \node[attrs,anchor=east,shift={(-.5mm,0mm)}] (E3S) at (E3.east) {$\mathsf{s}$};
    \node[attrs,anchor=east,shift={(-.5mm,0mm)}] (E4S) at (E4.east) {$\mathsf{s}$};
    \node[attrs,anchor=east,shift={(-.5mm,0mm)}] (E5S) at (E5.east) {$\mathsf{s}$};
    
    \node[attrs,anchor=east,shift={(-.5mm,0mm)}] (P1S) at (P1.east) {$\mathsf{s}$};
    \node[attrs,anchor=east,shift={(-.5mm,0mm)}] (P2S) at (P2.east) {$\mathsf{s}$};
    \node[attrs,anchor=east,shift={(-.5mm,0mm)}] (P3S) at (P3.east) {$\mathsf{s}$};
    \node[attrs,anchor=east,shift={(-.5mm,0mm)}] (P4S) at (P4.east) {$\mathsf{s}$};
    \node[attrs,anchor=east,shift={(-.5mm,0mm)}] (P5S) at (P5.east) {$\mathsf{s}$};

    \node[attrs,anchor=east,shift={(-5.2mm,0mm)}] (B1B) at (B1.east) {$\mathsf{b}$};
    \node[attrs,anchor=east,shift={(-5.2mm,0mm)}] (B2B) at (B2.east) {$\mathsf{b}$};

    \node[attrs,anchor=east,shift={(-.5mm,0mm)}] (B1R) at (B1.east) {$\mathsf{r}$};
    \node[attrs,anchor=east,shift={(-.5mm,0mm)}] (B2R) at (B2.east) {$\mathsf{r}$};
    
    \draw[-] (E1) -- (P1);
    \draw[-] (E2) -- (P1);
    \draw[-] (E2) -- (P2);
    \draw[-] (E2) -- (P3);
    \draw[-] (E3) -- (P3);
    \draw[-] (E4) -- (P3);
    \draw[-] (E4) -- (P4);
    \draw[-] (E5) -- (P4);
    \draw[-] (E5) -- (P5);

    \draw[-] (P1) -- (B1);
    \draw[-] (P2) -- (B1);
    \draw[-] (P3) -- (B2);
    \draw[-] (P4) -- (B2);
    \draw[-] (P5) -- (B2);
    \end{tikzpicture}
    \caption{Relational Skeleton $\sigma\in\Sigma_{\schema}$}
    \label{fig:skeleton}
\end{subfigure}%
\hfill\null%

\hfill%
\begin{subfigure}{0.4\textwidth}\centering
    \begin{tikzpicture}
    \draw[-,opacity=0] (0,-1.4) -- (0,0.7);
    \begin{scope}[opacity=0.2]
        \node[attr,opacity=1] (EC) {\textsf{competence\vphantom{fg}}};
        \node[attr,opacity=1] (ES) [below=1mm of EC] {\textsf{salary\vphantom{fg}}};
        \node[entc,fit=(EC)(ES), draw,label=above:{\textsf{Employee}}] (E) {};

        \node[attr,opacity=1] (BR) [right=2.5cm of EC] {\textsf{revenue\vphantom{fg}}};
        \node[attr,opacity=1] (BB) [below=1mm of BR] {\textsf{budget\vphantom{fg}}};
        \node[entc,fit=(BR)(BB), draw,label=above:{\textsf{Biz-Unit}}] (B) {};
        
        \node[attr,opacity=1] (PS) at ($(E.east)!0.5!(B.west)$){\textsf{success\vphantom{fg}}};
        \node[entc,fit=(PS), draw,label=above:{\textsf{Product}}] (P) {};
        
        \node[relc,label=left:{\textsf{Develops}}] (D) [below left=5mm and 1mm of P] {};
        \node[relc,label=right:{\textsf{Funds}}] (F) [below right=5mm and 1mm of P] {};

        \draw[-] (D) -- (E) node[midway,label=left:{\textsf{\tiny m}}] {};
        \draw[-] (D) -- (P) node[pos=0.8,label=right:{\textsf{\tiny m}}] {};
        \draw[-] (F) -- (P);
        \draw[-] (F) -- (B) node[midway,label=right:{\textsf{\tiny m}}] {};
    \end{scope}
    \draw[thick,->,cBe] (EC.west) to [bend right=90,looseness=1.5] (ES.west);
    \draw[thick,->,cBe] (BR.10) to [bend left=60,looseness=1.5] (BB.-10);
    \draw[thick,->,rounded corners=2mm,cBe] (EC.-30) -- (D.north) -- (PS.210);
    \draw[thick,->,rounded corners=2mm,cBe] (PS) -- (F.north) -- (BR.200);
    \draw[thick,->,rounded corners=2mm,cBe] (BB) -- ($(F.center)+(0,-1mm)$) -- (P.south) -- ($(D.center)+(0,-1mm)$) -- (ES);
    \end{tikzpicture}
    \caption{Relational Causal Model $\MM$}
    \label{fig:rcm}
\end{subfigure}\hfill
\begin{subfigure}{0.5\textwidth}\centering
    \begin{tikzpicture}[node distance=3.5mm]
    \draw[-,opacity=0] (0,-1.8) -- (0,.3);
    \begin{scope}[opacity=0.2]
    \node[skit] (E1) {$\mathsf{e_1}$};
    \node[skit] (E2) [right=of E1] {$\mathsf{e_2}$};
    \node[skit] (E3) [right=of E2] {$\mathsf{e_3}$};
    \node[skit] (E4) [right=of E3] {$\mathsf{e_4}$};
    \node[skit] (E5) [right=of E4] {$\mathsf{e_5}$};

    \node[skit] (P1) [below=3mm of E1] {$\mathsf{p_1}$};
    \node[skit] (P2) [right=of P1]     {$\mathsf{p_2}$};
    \node[skit] (P3) [right=of P2]     {$\mathsf{p_3}$};
    \node[skit] (P4) [right=of P3]     {$\mathsf{p_4}$};
    \node[skit] (P5) [right=of P4]     {$\mathsf{p_5}$};

    \node[skit] (B1) [below=3mm of P2] {$\mathsf{b_1}$};
    \node[skit] (B2) [below=3mm of P4] {$\mathsf{b_2}$};

    \draw[-] (E1) -- (P1);
    \draw[-] (E2) -- (P1);
    \draw[-] (E2) -- (P2);
    \draw[-] (E2) -- (P3);
    \draw[-] (E3) -- (P3);
    \draw[-] (E4) -- (P3);
    \draw[-] (E4) -- (P4);
    \draw[-] (E5) -- (P4);
    \draw[-] (E5) -- (P5);

    \draw[-] (P1) -- (B1);
    \draw[-] (P2) -- (B1);
    \draw[-] (P3) -- (B2);
    \draw[-] (P4) -- (B2);
    \draw[-] (P5) -- (B2);
    \end{scope}

    \node[attrs,anchor=east,shift={(-5.2mm,0mm)}] (E1C) at (E1.east) {$\mathsf{C}$};
    \node[attrs,anchor=east,shift={(-5.2mm,0mm)}] (E2C) at (E2.east) {$\mathsf{C}$};
    \node[attrs,anchor=east,shift={(-5.2mm,0mm)}] (E3C) at (E3.east) {$\mathsf{C}$};
    \node[attrs,anchor=east,shift={(-5.2mm,0mm)}] (E4C) at (E4.east) {$\mathsf{C}$};
    \node[attrs,anchor=east,shift={(-5.2mm,0mm)}] (E5C) at (E5.east) {$\mathsf{C}$};

    \node[attrs,anchor=east,shift={(-.5mm,0mm)}] (E1S) at (E1.east) {$\mathsf{S}$};
    \node[attrs,anchor=east,shift={(-.5mm,0mm)}] (E2S) at (E2.east) {$\mathsf{S}$};
    \node[attrs,anchor=east,shift={(-.5mm,0mm)}] (E3S) at (E3.east) {$\mathsf{S}$};
    \node[attrs,anchor=east,shift={(-.5mm,0mm)}] (E4S) at (E4.east) {$\mathsf{S}$};
    \node[attrs,anchor=east,shift={(-.5mm,0mm)}] (E5S) at (E5.east) {$\mathsf{S}$};
    
    \node[attrs,anchor=east,shift={(-.5mm,0mm)}] (P1S) at (P1.east) {$\mathsf{S}$};
    \node[attrs,anchor=east,shift={(-.5mm,0mm)}] (P2S) at (P2.east) {$\mathsf{S}$};
    \node[attrs,anchor=east,shift={(-.5mm,0mm)}] (P3S) at (P3.east) {$\mathsf{S}$};
    \node[attrs,anchor=east,shift={(-.5mm,0mm)}] (P4S) at (P4.east) {$\mathsf{S}$};
    \node[attrs,anchor=east,shift={(-.5mm,0mm)}] (P5S) at (P5.east) {$\mathsf{S}$};

    \node[attrs,anchor=east,shift={(-5.2mm,0mm)}] (B1B) at (B1.east) {$\mathsf{B}$};
    \node[attrs,anchor=east,shift={(-5.2mm,0mm)}] (B2B) at (B2.east) {$\mathsf{B}$};

    \node[attrs,anchor=east,shift={(-.5mm,0mm)}] (B1R) at (B1.east) {$\mathsf{R}$};
    \node[attrs,anchor=east,shift={(-.5mm,0mm)}] (B2R) at (B2.east) {$\mathsf{R}$};

    \begin{scope}[cBe,->]
        \draw (E1C) -- (P1S);
        \draw (E2C) -- (P1S);
        \draw (E2C) -- (P2S);
        \draw (E2C) -- (P3S);
        \draw (E3C) -- (P3S);
        \draw (E4C) -- (P3S);
        \draw (E4C) -- (P4S);
        \draw (E5C) -- (P4S);
        \draw (E5C) -- (P5S);
        \draw (P1S) -- (B1R);
        \draw (P2S) -- (B1R);
        \draw (P3S) -- (B2R);
        \draw (P4S) -- (B2R);
        \draw (P5S) -- (B2R);
        \draw (B1R) -- (B1B);
        \draw (B2R) -- (B2B);
        \draw (E1C) -- (E1S);
        \draw (E2C) -- (E2S);
        \draw (E3C) -- (E3S);
        \draw (E4C) -- (E4S);
        \draw (E5C) -- (E5S);
        \draw (B1B) -- (E1S);
        \draw (B1B) to [bend right=20] (E2S);
        \draw (B2B) to [bend left=20] (E2S);
        \draw (B2B) -- (E3S);
        \draw (B2B) to [bend left=20] (E4S);
        \draw (B2B) -- (E5S);
    \end{scope}
    \end{tikzpicture}
    \caption{Ground Graph $\GGG$}
    \label{fig:gg}
\end{subfigure}%
\hfill\null
\caption{An example of a relational schema, relational skeleton, relational causal model, and ground graph.}
\label{fig:prelim-example-ab}
\end{figure*}
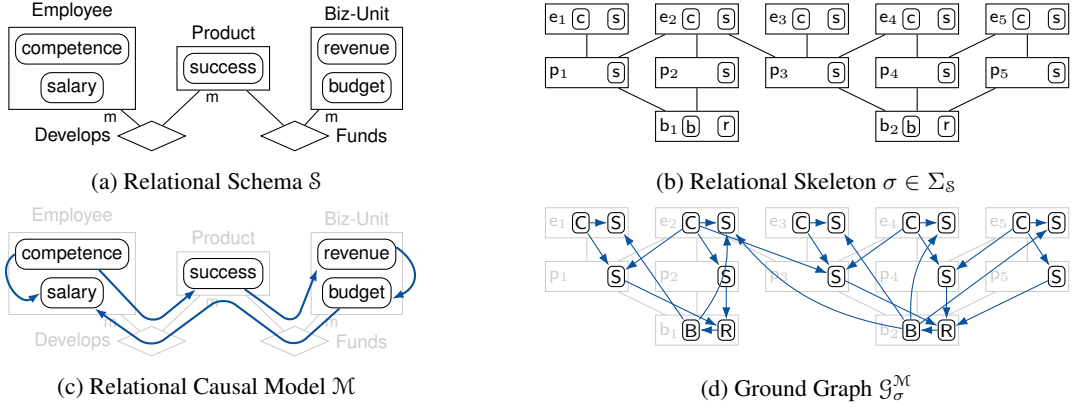

%% file: rpcd.tex

We first revisit RpCD \citep{Lee2016UAI} (See Appendix for the pseudocode), a sound and complete algorithm for learning the structure of a RCM from a given relational schema and access to a RCI oracle. RpCD was inspired by PC \citep{Spirtes2000book} for CBN, and RCD \citep{maier2013rcd} for RCM.
RpCD consists of two phases where the first phase identifies undirected RDs (i.e., adjacencies) based on answers to RCI queries and the second phase orients (a maximal subset of) the identified RDs based on answers to RCI queries and other known constraints.

\noindent
{\bf Phase I} Phase I of RpCD examines undirected RDs of the underlying model. 
The algorithm starts by enumerating a set of candidate RDs in an undirected form, which is analogous to preparing a complete (undirected) graph for CBN for further processing. Then, it removes undirected RDs of the form $P.Y-\VV{X}$ from the set of candidate RDs if a separating set $\bm{S}$ between $P.Y$ and $\VV{X}$ is found.

\noindent
{\bf Phase II} Phase II of RpCD orients a subset of identified undirected RDs based on the answers of queries posed to the RCI oracle.
Recall that, in the CBN literature, an undirected path $X- Y - Z$ where $X$ and $Z$ are not adjacent is called an {\em unshielded triple} (UT). The node $Y$ on the path $X \to Y \gets Z$ is called a collider. If $X$ and $Z$ are not adjacent, $Y$ is called an {\em unshielded collider}. The PC algorithm orients edges among vertices in an UT $X - Y - Z$ by finding a separating set $\bm{S}$ between $X$ and $Z$: $Y\notin \bm{S}$. 
\citet{Lee2016UAI} generalized the notion of UTs to the relational setting, and introduced \textit{canonical unshielded triple} (CUT for short) which has testable implications in the underlying RCM:
\begin{defn}[Canonical Unshielded Triple]
Let $\mathcal{M}$ be a RCM defined on a relational schema $\mathcal{S}$.
Suppose $\langle i.X,j.Y,k.Z\rangle$ is an unshielded triple in
the ground graph $\mathcal{G}_{\sigma}^{\mathcal{M}}$ for some $\sigma\in\Sigma_{\mathcal{S}}$.
There must be two (not necessarily distinct) dependencies $P.Y-\VV{X}$
and $Q.Z-\VV{Y}$ of $\mathcal{M}$ (ignoring directions)
such that $j\in P|_{i}^{\sigma}$ and $k\in Q|_{j}^{\sigma}$. Then,
we say that $\langle \VV{X},\,\bm{P}.Y,\,R.Z\rangle$ is a \emph{canonical
unshielded triple} (CUT) of $\mathcal{M}$ for every $R\in\left\{ T\mid k\in T|_{i}^{\sigma}\right\} $
where $\bm{P}=\{T\mid j\in T|_{i}^{\sigma}\}$.
\end{defn}
If a separating set is found for a CUT through answers RCI queries provided by a RCI oracle, then the edges between the relational variables in a CUT can be oriented in a RCM in a manner analogous to that of UTs in CBN. 
For example, consider an UT $\mathsf{\tuple{e_2.C,p_3.S,e_4.C}}$ in \Cref{fig:skeleton} based on a relational dependency $\mathsf{[PDE].C\to [P].S}$ (where $X=Z$). Then, the corresponding CUT is 
\begin{align}\label{eq:cut-edpde}
\mathsf{\tuple{[E].C, \set{[EDP].S}, [EDPDE].C}}.
\end{align}
A separating set $\bm{S}$ exists such that $\mathsf{[E].C} \Perp \mathsf{[EDPDE].C} \mid \bm{S}$, and $\bm{S}$ without $\mathsf{[EDP].S}$ indicates $\mathsf{[E].C} \to \mathsf{[EDP].S} \gets \mathsf{[EDPDE].C}$, that is, $\mathsf{[PDE].S}\to \mathsf{[E].C}$.

RpCD further orients a maximal subset of the rest of the undirected RDs using simple rules that are analogous to those used by PC to orient the edges of a CBN (do not introduce any new unshielded colliders or cycles).

\noindent
{\bf Challenges to be overcome}
While \citet{Lee2016UAI} showed that RpCD, when given access to a RCM oracle, is guaranteed to
yield a correct partially-directed RCM structure, there remain significant hurdles to be overcome before RpCD becomes useful in practice:
i) a RCI oracle needs to be replaced with a suitable, sufficiently reliable RCI test. Existing CI tests are either unsuitable for RCI or suffer from low power and hence inability to detect RCI;
ii) Even a well-designed RCI test may not be sufficiently reliable when applied to small samples.
Incorrect results of RCI tests at early during the execution of the structure learning algorithm may irrecoverably misguide the algorithm.
iii) A generic RCI test may fail to account for the specific characteristics of a given relational data, thereby yielding suboptimal results.
We address the first challenge in \Cref{sec:cit}, and the second and third challenges in \Cref{sec:rcd}.

%% file: cit.tex

\begin{table}\centering
\footnotesize
\begin{tabular}{@{}lrrr@{}} \toprule
                     & $\mathsf{[EDPFB].B}$     & $\mathsf{[E].C}$       & $\mathsf{[EDP].S}$\\ \midrule
$\mathsf{e_1}$       & $\mathsf{\set{b_1.b}}$       & $\mathsf{\set{e_1.c}}$ & $\mathsf{\set{p_1.s}}$ \\
$\mathsf{e_2}$       & $\mathsf{\set{b_1.b,b_2.b}}$ & $\mathsf{\set{e_2.c}}$ & $\mathsf{\set{p_1.s,p_2.s,p_3.s}}$ \\
$\mathsf{e_3}$       & $\mathsf{\set{b_2.b}}$       & $\mathsf{\set{e_3.c}}$ & $\mathsf{\set{p_3.s}}$ \\
$\mathsf{e_4}$       & $\mathsf{\set{b_2.b}}$       & $\mathsf{\set{e_4.c}}$ & $\mathsf{\set{p_3.s,p_4.s}}$ \\
$\mathsf{e_5}$       & $\mathsf{\set{b_2.b}}$       & $\mathsf{\set{e_5.c}}$ & $\mathsf{\set{p_4.s,p_5.s}}$ \\ \bottomrule
\end{tabular}
\caption{A flattened representation of relational data (\Cref{fig:skeleton}) with respect to a RCI query in \Cref{eq:rciexample}.}
\label{tab:flat}
\end{table}

We proceed to consider the implications of using an existing CI test designed for iid data (CI test for short) to reliably answer a RCI query in the context of relational causal discovery using RpCD. Recall that iid data are often stored in a single table where the columns correspond to the variables and rows are populated by (iid) instances. Suppose, given a RCI query, we were to flatten (or propositionalize) the relational data to obtain a RCI query specific single table as follows: To test $P.X\Perp Q.Y \mid R.Z$ against relational data, we create a table wherein each row corresponds to a \emph{base item} $i\in \sigma(B)$ of the common base item class $B$ of $P$, $Q$, and $R$ and the three columns of the table correspond to $P.X$, $Q.Y$ and $R.Z$ such that the cell for row $i$ and column $P.X$ is a multi-set $P.X|_i^\sigma$. Let us call the resulting data flattened data for short.
For example, \Cref{tab:flat} shows a table with three columns constructed to answer a RCI query (\Cref{eq:rciexample}) where the leftmost column corresponds to the row identifier.\footnote{If a cell for
$P.X$ or $Q.Y$ is empty, we discard the corresponding row from the table.}
It is not difficult to observe that the rows of the table constructed using the procedure described above are clearly not independent because, multiple rows of the table, e.g., $P.X|_i^\sigma$ and $P.X|_j^\sigma$, can share the same attributes. Needless to say, flattening does not get rid of the non-iid nature of relational data, which means that, in general, a CI test when applied to the table resulting from the flattening process may incorrectly reject the null hypothesis (independence) although $P.X\Perp Q.Y \mid R.Z$. In light of the preceding observation, are there conditions under which a CI test when applied to the table resulting from the RCI query specific flattening process described above is guaranteed to correctly determine whether or not RCI holds? To answer this question, we revisit the Relational Causal Markov Condition (RCMC, \citet{maier2014thesis}), which states that a canonical relational variable is independent of its non-descendants given its parents. We first recall the definition of non-descendants of a relational variable of a RCM before proceeding to revisit RCMC.

\begin{defn}
Let RCM $\MM$ be defined on a schema $\schema$, and 
let $W$ and $\VV{X}$ be different relational variables defined on $\schema$ sharing a common perspective $B$.
Then, $W$ is \emph{non-descendant}
of $\VV{X}$ if 
$W|_{b}^{\sigma}\cap de\left(b.X;\mathcal{G}_{\sigma}^{\MM}\right)=\emptyset$ for every $\sigma\in\Sigma_{\schema}$ and $b\in\sigma(B)$.
\end{defn}
\begin{defn}[Relational Causal Markov Condition]
Given a RCM $\MM$ defined on a relational schema $\schema$, 
$\bm{W}|_{b}^{\sigma} \Perp b.X \mid pa(b.X;\GGG)$
for every $X\in\bm{A}$, $\sigma\in\Sigma_{\schema}$, and $b\in\sigma(I_X)$
if $\bm{W}$ is a set of non-descendants of $\VV{X}$.
\end{defn}
RCMC implies that $\bm{W} \Perp \VV{X} \mid pa(\VV{X};\MM)$. A RCI query of the form $U\Perp \VV{X} \mid \bm{Z}$ is said to be RCMC-related
if $U$ is non-descendant of $\VV{X}$, and $\bm{Z}$ consists of the parents of $\VV{X}$ and does not include any non-descendant of $\VV{X}$.
We claim that any RCI query that is RCMC-related can be correctly answered using a CI test (where a random variable can assume values that are multi-sets) applied to the RCI query-specific flattened table constructed using the procedure described above. To see why this claim is true, note that given $pa(\VV{X};\MM)$, attributes of $\VV{X}$ must be independent and identically distributed, regardless of other conditioned non-descendants. Thus, the variability of $\VV{X}$ across different conditions arises from external factors that are independent of the non-descendants of $\VV{X}$. Hence, a traditional CI test applied to the flattened data can accurately answer a RCMC-related RCI query against relational data.
For example, consider a generic model where $\VV{X}\gets f(pa(\VV{X};\MM), \epsilon)$.
Given a fixed value for $pa(\VV{X};\MM)$, $\VV{X}$ can be viewed as $g(\epsilon)$ for some function $g$. Since $g(\epsilon)$ is independent of the non-descendants of $\VV{X}$, a CI test will correctly assert $\bm{W} \Perp \VV{X} \mid pa(\VV{X};\MM)$.

Note that although as we showed RCI which is RCMC-related can be correctly answered using a CI test, we can offer no such guarantee in the general case of a RCI query that is not RCMC-related. Specifically, in the general case, such a procedure can fail to establish RCI although RCI holds. Fortunately, however, the violation of iid assumption does not interfere with the CI test rejecting the null hypothesis (independence) when RCI does not in fact hold. Based on this understanding of the conditions under which a CI test can be used to reliably substitute for RCI tests against relational data, we can modify RpCD to substitute RCI oracle with a CI test applied to a RCI-query-specific flattening of the relational data.

%% file: rcd.tex

There has been much work on making causal discovery from iid data robust in the presence of limited data or violations of key assumptions \citep{Dash1999,Abellan2006,Ramsey2006,Cano2008,Bromberg2009}, including on methods that take advantage of recent advances in general-purpose Boolean satisfiability solvers \citep{Hyttinen2013cyclic,Triantafillou2015jmlr,magliacane2016nips}. Hence, in what follows, we focus our discussion primarily on approaches to making causal discovery robust that are specific to the relational (as contrasted with the iid) setting. We proceed to consider the two key phases of RpCD in turn.

\subsection{PHASE I: IDENTIFYING ADJACENCIES}
Recall that RpCD starts by initializing a set of candidate relational dependencies (RDs) given a user-specified maximum hop length of RDs to be considered. Let $\MM'$ be an intermediate RCM at an intermediate step during the execution of phase I of RpCD. In light of the results of the previous section regarding the conditions under which RCI tests against relational data can be reliably substituted by CI tests against an appropriate flattening of the relational data, we can ensure that RpCD first asks RCI queries that match RCMC to eliminate spurious candidate dependencies, while retaining the genuine dependencies. Recall that RpCD performs RCI tests to determine whether a candidate neighbors of a canonical relational variable (CRV) is in fact a genuine neighbor.
Since at the outset, the candidate neighbors of a CRV include the genuine parents of the CRV in the RCM, RpCD will eventually test
$\left(P.X\Perp \VV{Y}\mid pa\left(\VV{Y};\MM'\right)\right)$
for any connection $P.X-\VV{Y}$ unless it is disconnected with a separating set other than the parents of the CRV.
Note that any incorrect answers to the RCI queries (i.e., incorrect rejections of the null hypothesis (independence) by the CI tests) do not adversely impact the correctness of the algorithm unless the RCI query matches RCMC. However, there is the possibility of incorrectly discarding true RDs. 
For example, it is possible that $P.X\to \VV{Y}\in \bm{D}$ might be discarded from the candidate list due to relatively weak dependence. We proceed to examine a way to contain the deleterious impact of incorrectly discarding true dependencies.

\noindent
{\bf Order-independence} 
Whenever a true RD is discarded, it has a cascading deleterious effect on future steps of the PC algorithm (and its variants) which, as shown by \citet{Colombo2014order}, can however be avoided 
by making the necessary modifications to render the algorithms independent of the order in which variables are considered. 
Such modifications can be directly incorporated into RpCD: Prepare an empty set, store dependencies to be removed in the set instead of removing them immediately, and remove dependencies in the set when the algorithm proceeds to consider larger conditionals.

\noindent
{\bf Asymmetry and Aggregation}
There exists a notable difference between an edge in a CBN and a RD in a RCM with respect to the test for adjacencies ---
there can be two different tests for an adjacency.\footnote{\citet{Arbour2016UAI} considered asymmetry in relational data to infer the orientation of an underlying dependency while our focus is to improve the power of CI test.}
Through Phase I, we seek to ensure that the following holds true
$\parens{P.X \notperp \VV{Y} \mid \bm{W}}$
for $P.X\to \VV{Y}\in \bm{D}$ for a set of RVs $\WW$ with base item class $I_Y$ and $P.X\notin\WW$.
The test for the same adjacency must be performed from a different perspective $I_X$,
$\parens{\tilde{P}.Y \notperp \VV{X} \mid \bm{R}}$
where $\tilde{P}$ corresponds to $P$ reversed. Performing both tests is essential since the algorithm does not know in advance the topological order between $X$ and $Y$, and which RCI queries are in fact RCMC-related.

Consider 
a RD, $\mathsf{[PDE].Competence}\rightarrow\mathsf{[P].Success}$. The dependency between $\mathsf{[EDP].S}$
and $\mathsf{[E].C}$ can be substantially weaker
than the dependency between $\mathsf{[PDE].C}$ and $\mathsf{[P].S}$
since there is a set of coworkers whose competence affects
$\mathsf{[EDP].S}$ that is not considered.
For instance, 
$\mathsf{e}_2$ develops $\mathsf{p}_1$, $\mathsf{p}_2$, and $\mathsf{p}_3$ where the success of $\mathsf{p}_1$ and $\mathsf{p}_3$ are also determined by $\mathsf{e}_1$, and $\mathsf{e}_3$ and $\mathsf{e}_4$, respectively, diluting the strength of the relationship between competence of the worker(s) and the success of the product.


To protect against the possibility that a RCI test wrongly failing to reject independence, e.g., $(P.X\Perp \VV{Y} \mid \bm{W})$, we can perform an \emph{additional} test. We can first apply an aggregating function (e.g., mode, average, median, etc), 
on $P.X\in adj\left(\VV{Y};\MM'\right)$, and then 
conduct an additional test which (modulo slight abuse of notation) is given by
\[
\left(f\left(P.X\right)\Perp \VV{Y}\mid\bm{W}\right)_{\sigma}
\]
where each $P.X|_{i}^{\sigma}$ is replaced by $f\left(P.X|_{i}^{\sigma}\right)$.
Such aggregation does \emph{not} introduce spurious dependencies. However, in practice,
it can help overcome weak RCI tests as the mapping reduces
not only the dimensionality of the variables involved in the test but also the variances caused by exogenous variables.


It is worth noting that aggregation is widely used for dealing with RVs in relational machine learning \citep{perlich2006distribution} as well as relational causal discovery \citep{maier2013rcd}. However, an important distinction between their use of aggregation and ours is that we do not apply an aggregate function on the conditionals, for doing so may result in false positive answers to RCI queries.
For example, consider $X\to Z\to Y$ where $X\Perp Y \mid Z$. If we transform only $X$ using the aggregation function $f$, we get
$f(X) \gets X \to Z \to Y$ where $f(X) \Perp Y \mid Z$ still holds true. If $Z$ is transformed through the aggregation function $g$, we get $X\to Z \to Y$ with $Z \to g(Z)$, and, thus, $X \notperp Y \mid g(Z)$ nor $f(X) \notperp Y \mid g(Z)$.
Note that aggregation presented here also applies to Phase II of the algorithm described in the next subsection.

\subsection{PHASE II: ORIENTING RELATIONAL DEPENDENCIES}

A RCI test against a canonical unshielded triple (CUT) can establish the orientation of edges among the vertices in the triple. Since the purpose of the test is to find a separating set, false positives for non-RCMC-related queries are a non-issue.
However, weak dependence can lead to false negatives, and hence an invalid separating set, which may include colliders or may exclude non-colliders, resulting in an incorrect orientation of the edges among the vertices in a CUT.

Since RCM assumes acyclicity at an attribute class level, once we perform a RCI test on a CUT $\tuple{\VV{X},\bm{P}.Y,R.Z}$, assuming that the test is reliable, there is no need to test on other CUTs with matching attribute classes i.e., $\tuple{X,Y,Z}$ (or $\tuple{Z,Y,X}$). 
However, given the possibility of erroneous results from RCI tests, we can perform tests on multiple CUTs to determine the  orientation of an edge, e.g., $X\to Y\gets Z$, and make an informed decision by, e.g., majority rule, two-thirds, etc, based on the results of multiple tests. The algorithm may even  
obtain multiple separating sets against a CUT and check them for consistency (e.g., orientation-faithfulness \citep{Ramsey2006}). However, as we shall show below, care must be exercised in how these ideas are incorporated into RpCD.

\noindent
{\bf Limitations of CUT-based RCI tests} We start by examining why \emph{naively} conducting RCI tests against CUTs is not the best idea.
To see why, consider the flattening of the CUT in \Cref{eq:cut-edpde} for answering a RCI query using a CI test. Since $\mathsf{e}_2$ develops three products $\mathsf{\set{p_1,p_2,p_3}}$, $\mathsf{e}_2$'s coworkers are $\mathsf{\set{e_1,e_3,e_4}}$.
Success of $\mathsf{p}_2$ is, in fact, irrelevant to the competences of the coworkers $\set{\mathsf{e_1,e_3,e_4}}$, whereas the  success of each product depends only  on the competence of the employees who develop it (e.g.,  $\mathsf{e}_1$ in the case of $\mathsf{p}_1$). (i) Given a CUT $\tuple{\VV{X},\bm{P}.Y,R.Z}$ and some $P\in\bm{P}$, the association between $P.Y$ and $R.Z$ seems relatively weaker than that between two RVs connected by a RD.
(ii) Further, $R$ can be long, and the average dimensionality of $\set{R|_i^\sigma}_{i\in\sigma(I)}$ can be large. If each employee develops $k$ products, and each product is developed by $m$ employees, then, each employee can have up to $km-1$ coworkers.
(iii) Unlike the unshielded triples of a CBN, absence of a RD between $\VV{X}$ and $R.Z$ does not imply that there is no connection between $i.X$ and $R.Z|_i^\sigma$ for some item $i$. If $R'.Z\to \VV{X}\in\mathbf{D}$, then there will be edges from $i.X$ to $R.Z|_i^\sigma \cap R'.Z|_i^\sigma$.

In light of the preceding observations we consider alternative ways to perform RCI tests that are relevant to CUTs.
We can classify RCI tests for orientating the RDs into the following categories:  (i) relational bivariate orientation (RBO) which can be further subdivided into split-RBO and pair-RBO, and (ii) non-RBO.\footnote{RBO is proposed as a special rule in \citet{maier2013rcd} where a non-collider e.g., not $X\to Y\gets X$ implies $Y\to X$. We rather use the term RBO not as an ``orientation rule'' but as a way to categorize CUT-based RCI tests.} We proceed to discuss each of these in turn.

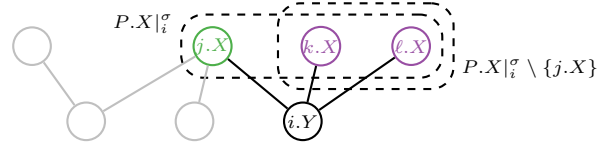
\begin{figure}\centering
\scriptsize
\begin{tikzpicture}[circ/.style={circle,draw,minimum size=5mm, inner sep=0mm}, node distance=5.mm,x=1.2cm,y=1.cm,thick]
    \node[circ] (d) {$i.Y$};
    \node[circ,gray!50] (e) at (-1.2,0) {};
    \node[circ,gray!50] (f) at (-2.4,0)  {};
    \node[circ,bettergreen] (a) at (-1,1) {$j.X$};
    \node[circ,betterpurple] (b) at (0.2,1)  {$k.X$};
    \node[circ,betterpurple] (c) at (1.2,1) {$\ell.X$};
    \node[circ,gray!50] (c2) at (-3,1) {};
    \node[rounded corners=3mm,rectangle,fit=(b)(c),draw,dashed,inner sep=3mm] (fit1) {};
    \node[rounded corners=3mm,rectangle,fit=(a)(b)(c),draw,dashed, inner sep=1.5mm] (fit2) {};
    \node[anchor=west,yshift=-3mm] at (fit1.east) {$P.X|_i^\sigma\setminus\set{j.X} $};
    \node[anchor=east,yshift=3mm] at (fit2.west) {$P.X|_i^\sigma$};
    \draw[-,gray!50] (a) -- (e);
    \draw[-,gray!50] (a) -- (f);
    \draw[-,gray!50] (c2) -- (f);
    \draw[-] (a) -- (d);
    \draw[-] (b) -- (d);
    \draw[-] (c) -- (d);
    
\end{tikzpicture}
\caption{An instance in row-$j$ for split-RBO where three variables are split so as to create three rows. CUT-based tests will include gray vertices.}
\label{fig:split-rbo}
\end{figure}

\noindent{\bf Split-RBO} Relational Bivariate Orientation (RBO) is an orientation for a CUT where the two ends share the same attribute class. Given an undirected RD $P.X-\VV{Y}$ where $P$ is of  cardinality `many',
a CUT can be made $\tuple{\VV{X},\tilde{\bm{P}}.Y, R.X}$ where one can orient $X\to Y$ if it turns out to be a collider or $X\gets Y$, otherwise.
See \Cref{fig:split-rbo} where an item attribute $i.Y$ has three neighbors for $P.X|_{i}^\sigma$ in an intermediate undirected ground graph.
From the perspective of $i.Y$, $P.X|_{i}^\sigma$ can be split into two parts: a singleton (one) and the rest of the item attributes (rest). Then, a separating set will be obtained from a subset of neighbors of $\VV{X}$ in an intermediate model. The corresponding test can be expressed as
\begin{align}\label{eq:split}
\mathsf{one}(P.X) \Perp \mathsf{rest}(P.X) \mid \bm{S}
\end{align}
where $\bm{S}\subseteq adj(\VV{X};\MM')$.
To carry out the test, we construct a flattened representation as follows.
Among $\sigma(I_Y)$, find items that are connected to at least two $I_X$ items through $P$ in $\sigma$.
For each element $j.X \in P.X|_i^\sigma$, create a row
$\tuple{j.X, P.X|_i^\sigma\setminus\{j.X\}}$.
Then, refine the resulting table to ensure that the $j.X$ column is made of unique item attributes. Columns for $\bm{S}$ can be added later.


\begin{table}\centering
\footnotesize
\begin{tabular}{@{}lrrr@{}} \toprule
               & $\mathsf{[PDE].C}$                 & $\mathsf{[PDE].C\setminus e_i.C}$                       & $\mathsf{e_i.C}$\\ \midrule
$\mathsf{p_1}$ & $\mathsf{\set{e_1.c, e_2.c}}$      & $\mathsf{\set{\textcolor{gray!50}{e_1.c}, e_2.c}}$      & $\mathsf{e_1.c}$\\
               &                                    & $\mathsf{\set{e_1.c, \textcolor{gray!50}{e_2.c}}}$      & $\mathsf{e_2.c}$\\ [1ex]
$\mathsf{p_3}$ & $\mathsf{\set{e_2.c,e_3.c,e_4.c}}$ & $\mathsf{\set{\textcolor{gray!50}{e_2.c},e_3.c,e_4.c}}$ & $\mathsf{e_2.c}$\\
               &                                    & $\mathsf{\set{e_2.c,\textcolor{gray!50}{e_3.c},e_4.c}}$ & $\mathsf{e_3.c}$\\
               &                                    & $\mathsf{\set{e_2.c,e_3.c,\textcolor{gray!50}{e_4.c}}}$ & $\mathsf{e_4.c}$\\ [1ex]
$\mathsf{p_4}$ & $\mathsf{\set{e_4.c,e_5.c}}$       & $\mathsf{\set{\textcolor{gray!50}{e_4.c},e_5.c}}$       & $\mathsf{e_4.c}$\\
               &                                    & $\mathsf{\set{e_4.c,\textcolor{gray!50}{e_5.c} }}$      & $\mathsf{e_5.c}$\\ \bottomrule
\end{tabular}
\caption{Flattened representation (before deduplication) for split-RBO test ($\mathsf{[PDE].C}$ for reference)}
\label{tab:splitrbo}
\end{table}

As an example, consider a representation to be used for split-RBO with respect to $\mathsf{[EDP].S - [P].C}$ (\Cref{tab:splitrbo}). Note that there are multiple employees connected to a product. For each such product, we can list product-specific coworkers who can be split into a singleton and the rest.
One might wonder if we can get by with using an employee and only one of his/her coworkers.
Consider \Cref{fig:split-rbo} again where $P.X$ is the cause of $\VV{Y}$. Assume, for simplicity, $i.Y=\sum_{j.X\in P.X|_i^\sigma} j.x$. If we were to consider only two singletons instead of one singleton and the rest, the relationship between the two singletons with respect to their common effect will become low. 

\noindent{\bf Pair-RBO} Since split-RBO already covered a case where $P$ is of cardinality `many', 
we now consider a type of RBO with two different candidate undirected RDs $P.X-\VV{Y}$ and $Q.X-\VV{Y}$ where both $P$ and $Q$ are of  cardinality `one'.
Due to their cardinalities, $P$ and $Q$ are not intersectable, and  every item attribute $j.Y$ for $j\in \sigma(I_Y)$, there can be at most two item attributes of $X$ representing $P.X$ and $Q.X$, i.e., $\verts{P.X|_i^\sigma}\leq 1$ and $\verts{Q.X|_i^\sigma}\leq 1$.
Hence, a separating set is obtained for pairs of singletons. If multiple RDs are defined for two attribute classes $X$ and $Y$, multiple pair-RBO and split-RBO tests can be used to orient them. Techniques described in \citet{Colombo2014order} for handling conflicting orientations in a CBN can be adopted to RCM in a relatively straightforward manner.

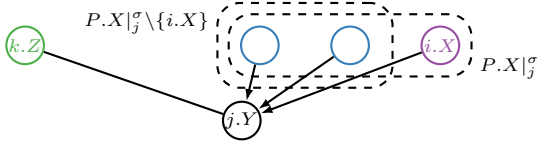
\begin{figure}\centering
\scriptsize
\begin{tikzpicture}[circ/.style={circle,draw,minimum size=5mm, inner sep=0mm}, node distance=5.mm,x=1.2cm,y=1.cm,thick]
    \node[circ] (d) {$j.Y$};
    \node[circ,bettergreen] (a) at (-2.4,1) {$k.Z$};
    \node[circ,betterblue] (b) at (0.2,1)  {};
    \node[circ,betterblue] (c) at (1.2,1) {};
    \node[circ,betterpurple] (c1) at (2.2,1) {$i.X$};
    \node[rounded corners=3mm,rectangle,fit=(b)(c),draw,dashed,inner sep=3mm] (fit1) {};
    \node[rounded corners=3mm,rectangle,fit=(b)(c)(c1),draw,dashed, inner sep=1.5mm] (fit2) {};
    \node[anchor=east,yshift=3mm] at (fit1.west) {$P.X|_j^\sigma{\setminus}\set{i.X} $};
    \node[anchor=west,yshift=-3mm] at (fit2.east) {$P.X|_j^\sigma$};
    \draw[-] (a) -- (d);
    \draw[->] (b) -- (d);
    \draw[->] (c) -- (d);
    \draw[->] (c1) -- (d);
\end{tikzpicture}
\caption{Non-RBO with $P.X{\to} \VV{Y}$ and $Q.Z{-}\VV{Y}$}
\label{fig:non-split-rbo}
\end{figure}

\noindent{\bf Non-RBO} Now, we seek for a principled approach to orienting RDs when three different attribute classes are involved in a CUT, say $\tuple{\VV{X},\tilde{\bm{P}}.Y,R.Z}$ with $P.X-\VV{Y}$ and $Q.Z-\VV{Y}$ such that $X\neq Z$. Performing RBO tests before performing any non-RBO tests offers several advantages:
(i) If RBO-tests between $X$ and $Y$ fail to orient a RD, we can conclude that $X$ and $Y$ exhibit weak dependence in relational data, in which case, it is advisable to  avoid  non-RBO tests that involve $X$ and $Y$, which may result in incorrect orientations; and
(ii) If some of the RDs are oriented (e.g., $X\prec Z$), then we can limit the tests to be performed to those that seek a separating set only from $\VV{Z}$. Further, a separating set $\bm{S}$ will be $pa(\VV{Z};\MM')\subseteq \bm{S}\subseteq adj(\VV{Z};\MM')$, which can reduce the number of tests needed to obtain a separating set.
(iii) At least one of $P$ and $Q$ is of  cardinality `one', permitting us to use a CI test (on suitably flattened data) in place of non-RBO RCI tests. If both are of cardinality `one', non-RBO tests can be done in a manner similar to pair-RBO.
(iv) Finally, if $Q$ be of cardinality  `one', then the orientation between $X$ and $Y$ would have already been determined (because a split-RBO test, it would precede a non-RBO test). Further, if $\tilde{P}.Y\to \VV{X}$ then, no RCI test is required since the CUT is a non-collider.
Hence, the only case left for non-RBO test is $P.X\to \VV{Y} - Q.Z\in\MM'$ with $P$ being of  cardinality `many'. Without knowing whether $X\not\prec Z$ nor whether $Z\not\prec X$, we may need to examine separating sets from both $I_X$ and $I_Z$ perspectives. First, from $I_Z$ perspective, RCI test can be performed in a table where row-$j$ for $j\in \sigma(I_Y)$ represents
$\tuple{k.Z, P.X|_j^\sigma, \dotsc, S^\ell|_k^\sigma,\dotsc }$
where $\set{k}=Q|^\sigma_j$ and $pa(\VV{Z};\MM')\subseteq\bm{S}\subseteq adj(\VV{Z};\MM')$.
With $I_X$ perspective, we should pick one of $P.X|_j^\sigma$, i.e., $\mathsf{one}(P.X)$, for row-$j$ and test against $Q.Z|_j^\sigma$, which is a singleton. Similarly, a separating set $\bm{S}$ satisfies  $pa(\VV{X};\MM')\subseteq\bm{S}\subseteq adj(\VV{X};\MM')$.

\noindent
{\bf Detecting Weak Dependence} \citet{Ramsey2006} explored techniques for determining whether CI tests against the given data yield consistent  orientations of  edges in a CBN.
Given an unshielded triple $X-Y-Z$, one can examine whether $Y$ appears in every $\bm{S}\subseteq adj(\{X,Z\};\GG)$ such that $X\Perp Z\mid \bm{S}$ and 
not in $\bm{W}\subseteq adj(\{X,Z\};\GG)$ such that $X\notperp Z\mid \bm{W}$. However, because
this not only is time consuming and but also yields very conservative results, a more pragmatic alternative is to examine
whether both $(X\Perp Z\mid \bm{S})$ and $(X\Perp Z\mid \bm{S}\cup\{Y\})$ result independence for some separating set $\bm{S}$ which does not include $Y$.
The idea can be incorporated  into RpCD where one can check, for example, with conditionals $\bm{S}\cup \set{\tilde{P}.Y}$ in testing against a CUT $\tuple{\VV{X},\tilde{\bm{P}}.Y, Q.Z}$ where $\tilde{P}\in \tilde{\bm{P}}$. 
In contrast, given our approaches (i.e., split-, pair-, and non-RBO), a better test can be achievable by limiting the item attributes of $Y$ to only those that are relevant to $P.X$, which is $\VV{Y}$. For split-RBO with a separating set $\bm{S}$, the following test can be performed, 
\[\mathsf{one}(P.X) \Perp \mathsf{rest}(P.X) \mid \bm{S}\cup \set{\VV{Y}}.\]
This detection mechanism can be applied to pair-RBO and non-RBO cases.

%% file: experiments.tex
\noindent
{\bf Experiments} We conducted experiments with synthetic data generated from known RCMs to assess how the proposed approaches
to replacing RCI oracle with RCI tests against relational data impact the performance of RpCD. In our implementation, we used two kernel-based CI tests, HSIC \citep{Gretton2005} for marginal, and SDCIT \citep{LH2017SDCIT} for conditional RCI queries. In our kernel-based CI test for multi-set valued random variables, we used, following \citet{Haussler1999}, $K'(\xx,\yy;\theta)=\sum_{x\in\xx}\sum_{y\in\yy}K(x,y;\theta)$.
In the case of real-valued data, $K$ is chosen to be a RBF kernel whose parameter $\theta$ is chosen using the median heuristic \citep{Gretton2007}. The resulting kernel matrices are normalized, e.g., $\bm{K}_{a,b}\gets \bm{K}_{a,b}/\sqrt{\bm{K}_{a,a}K_{b,b}}$.

We tested the performance of RpCD with the improvements proposed in this paper and the baseline (RpCD using the CI test for RCI with no other changes) on 300 randomly generated RCMs of varying complexity. For each RCM, we randomly generated different sizes of relational data with $n=200$ to $500$ resulting in approximately $n$ items per entity class and $2n$ relationships per relationship class. We parametrized the RCM using an adaptation of a linear Gaussian model to a relational setting. We used Average as the aggregation function. Additional details about the experimental setup and results are provided in the Appendix.\footnote{Code is available at \url{https://github.com/sanghack81/RRCD}}

\begin{table}
\footnotesize
\centering
\begin{tabular}{@{}ccccc@{}}
\toprule
Aggregation & Order Ind. & Base size & Precision & Recall \\ \midrule
False & False & 200 & 98.23 & 61.91 \\
      &       & 500 & 98.90 & 77.92 \\
      & True  & 200 & 98.86 & 57.95 \\
      &       & 500 & 99.13 & 76.77 \\ 
True  & False & 200 & 97.75 & 65.44 \\
      &       & 500 & 98.87 & 80.31 \\
      & True  & 200 & 98.75 & 61.77 \\
      &       & 500 & 99.09 & 79.03 \\ \bottomrule
\end{tabular}
\caption{Precision and recall (based on macro-average) for discovering undirected RDs in Phase I.}
\label{tab:p1:main}
\end{table}

\noindent
{\bf Phase I Experimental Results}
We find that (see \Cref{tab:p1:main}), as the size of relational data increases, the performance of Phase I improves as expected. Order-independence mitigates the effect of early false negative RCI test results, perhaps at the expense of a slightly reduced recall of undirected relational dependencies. Aggregation improves the power of the test at the expense of a slight increase in the false positive rate of the test. Note that the high precision and relatively low recall implies that errors of RCI tests are mainly false negatives.

We additionally investigated the types of queries where relational conditional independence is correctly found to not hold  by the additional aggregation-based tests. Aggregation was especially effective in reducing the false negative rate of the tests of independence between a canonical RV and its child (tests in a reverse direction) while rarely producing false positives (see Appendix).

%
%
%

\begin{table}
\footnotesize
\centering
\begin{tabular}{@{}lcccccc@{}}
\toprule
 & \multicolumn{3}{c}{200} &\multicolumn{3}{c}{500}\\ \cmidrule{2-7}
             & Acc. & C    & NC   & Acc. & C    & NC \\ \midrule
CUT (RBO)    & 54.0 & 46.1 & 60.7 & 54.1 & 46.4 & 60.6\\
P+S          & 68.7 & 64.7 & 72.6 & 75.6 & 73.4 & 77.9\\
w/ detection & 71.2 & 68.4 & 74.0 & 77.4 & 74.7 & 80.0 \\ \midrule
CUT          & 65.5 & 77.0 & 61.4 & 74.1 & 77.8 & 72.8 \\
P+S+N        & 74.0 & 73.2 & 74.7 & 80.4 & 78.2 & 82.2\\
w/ detection & 80.9 & 75.0 & 85.4 & 85.7 & 82.6 & 88.2\\ \bottomrule
\end{tabular}
\caption{Accuracies for orientation tests (overall (Acc.), collider (C), and non-colliders (NC)), for CUT-based tests, proposed tests, and with the weak dependency detection mechanism enabled. P, S, and N stands for Pair-RBO, Split-RBO, and Non-RBO, respectively.}
\label{tab:phase2rbo}
\end{table}

\noindent
{\bf Phase II Experimental Results}
Given the correct set of dependencies (which correspond to perfect Phase I results), we first performed experiments to measure the effectiveness of split-, pair-, and non-RBO tests relative to the CUT-based tests.
\Cref{tab:phase2rbo} shows the performance based on the first smallest separating set found. CUT-based tests do not perform well even with larger relational data (which also makes relational structure more complicated) for RBO cases. Proposed RBO tests outperform CUT-based tests regardless of the type of orientations, colliders or non-colliders, and show improvements with larger data. Additional weak dependency detection mechanism helps refining false negatives. A similar trend is observed when we also considered non-RBO cases. Note that, unlike Phase I, false negatives in Phase II might cause wrong orientations, thus, affecting both precision and recall.

Next, we compared our approach against a naive CUT-based approach by measuring the average precision and recall for the final orientations with respect to the true CPRCM\footnote{A CPRCM is a maximally-oriented RCM, which represents the Markov equivalence class of a RCM.} instead of the RCM. The final orientations for our approach were determined as follows: i) a majority vote rule is used to determine the orientation of each attribute class triple  (local)\footnote{Once a separating set of size $k$ is found, then we further examine whether there are other separating sets with size $k$, while a naive approach only makes use of RCI results based on the first separating set found.}; ii) the maximal non-conflicting local orientations are obtained (global). The baseline with CUT-based tests used the same majority rule but each orientation is accepted in a sequential manner if it does not cause conflicts with the already accepted orientations.
Given the perfect Phase I results, the precision and recall for Phase II based on our approach are 93.5\% and 75.4\%, respectively ($n=500$), as compared to 75.3\% and 69.7\%, respectively, for the CUT-based Phase II. Thus, we see substantial improvements over the baseline.

%% file: conclusions.tex

We introduced a robust algorithm for learning the structure of a
relational causal model from the given relational data.
We showed how a  conditional independence test designed for iid data can be used to 
effectively test for relational conditional independence  against relational data.
The relational causal Markov condition, a relational variable being independent of its non-descendants given its direct causes, allows the test to correctly establish relational conditional independence, whereas
the non-iid-ness of relational data helps the test to reject independence when independence does not hold. We introduced  several techniques to improve the robustness of the algorithm, and  empirically demonstrated their effectiveness. Despite these promising results, there is much room for further improvement, through better methods for testing independence of variables whose  values are multi-sets,  kernels optimized for the given relational data, as well as improved tests for relational conditional independence.

\subsubsection*{Acknowledgements} This work was funded in part by grants from the  NIH NCATS through the grant UL1 TR002014 and by the NSF through the grants 1518732, 1640834, and 1636795, the Edward Frymoyer Endowed Professorship  at Pennsylvania State and the Sudha Murty Distinguished Visiting Chair in Neurocomputing and Data Science funded by the Pratiksha Trust at the Indian Institute of Science (both held by Vasant Honavar). The content is solely the responsibility of the authors and does not necessarily represent the official views of the sponsors.

%% file: appendix.tex
\subsection*{RPCD ALGORITHM}
We present RpCD algorithm in \Cref{alg:rpcd}.
\begin{algorithm}[h]
\footnotesize
	\textbf{Input}: $\mathcal{S}$ relational schema, $h$ hop threshold
	\begin{algorithmic}[1]
		\State initialize $\bm{D}$ with candidate RDs up to
		$h$ hops.
		\State initialize an undirected graph $\MM' $ with undirected
		$\bm{D}$.
		\State $\ell\gets 0$
		\Repeat 
		\For{$(P.Y, \VV{X})$ \textbf{s.t.}
		$P.Y - \VV{X} \in \MM' $}
		\For{every $\bm{S}\subseteq ne(\VV{X};\MM' )\setminus\{P.Y\}$
		\textbf{s.t.} $\left|\bm{S}\right|=\ell$}
		\If{$P.Y \Perp \VV{X}\mid\bm{S}$ }
		\State remove $\{P.Y - \VV{X},\tilde{P}.X - \VV{Y}\}$
		from $\MM' $.
		\State \textbf{break}
		\EndIf
		\EndFor 
		\EndFor 
		\State $\ell\gets \ell+1$
		\Until{$|ne(\VV{X};\MM' )|-1<\ell$ for every $X\in\bm{A}$
		}\medskip{}
		\State initialize $\mathcal{U}$ with CUTs
		from $\MM' $.
		\State $\mathcal{N}\gets \emptyset$, $\mathcal{H}\gets \langle\bm{A},\{X-Y\mid P.Y-\VV{X}\in\MM' \}\rangle$
		\For{every $\langle \VV{X},\bm{P}.Y,R.Z\rangle\in\mathcal{U}$}
		\If{$\langle X,Y,Z\rangle \in \mathcal{N}$ \textbf{or} $\{X,Z\}\cap ne(Y;\mathcal{H})=\emptyset$ \textbf{or} $\{X,Z\}\cap ch(Y;\mathcal{H})\neq\emptyset$}
		\State \textbf{continue}
		\EndIf
		\If{exists $\bm{S} \subseteq adj(\VV{X};\MM' )$
		\textbf{s.t.} $R.Z \Perp \VV{X}\mid\bm{S}$}
		\If{$\bm{S}\cap\bm{P}.Y=\emptyset$}  orient $X \to Y \gets Z$
		in $\mathcal{H}$
		\ElsIf{$X = Z$} orient $Y \to X$ in $\mathcal{H}$
		\Else{ } add $\langle X,Y,Z \rangle$ to $\mathcal{N}$
		\EndIf
		\EndIf
		\State orient edges in $\mathcal{H}$ with sound rules with $\mathcal{N}$.
		\EndFor \medskip{}
		\State $completes\left(\mathcal{H},\mathcal{N}\right)$
		\State \textbf{return} $\bigcup_{P.Y-\VV{X}\in\MM' }\begin{cases}
		P.Y \to \VV{X} & Y\to X\in\mathcal{H}\\
		P.Y - \VV{X} & Y-X\in\mathcal{H}
		\end{cases}$ 
	\end{algorithmic}

	\caption{RpCD \citep{Lee2016UAI}}
	\label{alg:rpcd}
\end{algorithm}

\subsection*{RELATIONAL DATA}
We randomly generated 300 relational schemas of 3 (50\%), 4 (25\%), and
5 (25\%) entity classes with specified probabilities. Two to five relationship
classes are randomly generated to connect a pair (i.e., binary relationship)
of entity classes or a triple with 75\% and 25\% probability, respectively.
Cardinalities are selected uniformly. One to three attribute classes
are generated for each entity class, and zero or one attribute class
is generated for each relationship class. Finally, created relational
schemas that do not satisfy following rules are excluded: (i) all item
classes are connected and (ii) the total number of attribute classes are
less than or equal to 8.

RCMs are also generated randomly. Given a relational schema $\mathcal{S}$,
max hop length $h$ is selected uniformly between 2 to 4. The number
of dependencies is determined by $\lfloor\frac{3\left|\mathbf{A}\right|}{2}\rfloor$
and uniformly selected among \emph{all} relational dependencies within
the given $h$. We limit the maximum number of parents of a canonical
relational variable by 3. We reject generated RCMs if there exists
an isolated attribute class that does not involve any relational dependency. Further, if the CPRCM (a maximally-oriented RCM representing the Markov equivalence class of a RCM) of the generated
RCM has no directed dependencies, that is, the orientation of relational dependencies is impossible in theory. 
We adopt a linear model with additive
Gaussian noise using average aggregators: 
\[
i.X\gets \sum_{P.Y\in pa(\VV{X};\mathcal{M})}\frac{\beta_{P.Y,\VV{X}}}{|P.Y|_{i}^{\sigma}|}\parens[\Bigg]{\sum_{j.Y\in P.Y|_{i}^{\sigma}}j.Y}+\epsilon
\]
where $\beta_{P.Y,\VV{X}}=1+\left|\gamma\right|$ where $\gamma\sim\mathcal{N}\left(0,0.1^{2}\right)$
for every $P.Y\in pa\left(\VV{X};\mathcal{M}\right)$ for every $X\in\mathbf{A}$.
$\epsilon\sim\mathcal{N}\left(0,0.1^{2}\right)$. The set of parameters
will likely yield a relational data less hostile for our learning
algorithm given that $\beta\geq1$ and the variance of noise is relatively small.
This fulfills our intention to assess the behavior of learning algorithm
across different settings. If we wanted to exploit the fact that the
generated RCMs are based on an average aggregator, we could incorporate
this into the choice of kernel so that R-convolution kernel is not
necessary but a simple RBF kernel on averaged values is sufficient. 

Random relational skeletons are generated with a user-specified \emph{base}
size $n$. Given $n$, the number of relationships (i.e., relationship
instances or relationship items) for each relationship class is the
twice of the base size if the cardinality is `many' for every
its participating entity class and the same as base size, otherwise.
The number of entities per entity class can be computed as $\lfloor1.2^{k}\cdot n\rfloor$
where $k$ is the number of related relationship classes with all-`one'
cardinalities. 
For each RCM, we
generate 4 relational skeletons corresponding to base size from 200
to 500, increased by 100.

\subsection*{ROBUSTIFICATION of RPCD VS. NAIVE RPCD}

For the robust RpCD, we adopted features mentioned in the main paper: order-independence for Phase-I, and split-RBO, pair-RBO, non-RBO tests, and weak dependence detection for Phase-II. Aggregation-based additional tests are applied to both phases.
Separating sets are sought from the smallest size of conditionals to the largest. If a separating set is found with size $k$, the algorithm checks the existence of other separating sets of the same size, which we call `minimal separating sets'. Then, the orientation of relational dependencies is based on a majority vote for the orientation of each pair of attribute classes. At the end of the algorithm, different orientations are combined to yield a partially-oriented RCM (PRCM) which maximally satisfies obtained test results. If there are multiple candidate PRCMs matching the same number of test results, then we choose a PRCM, which has the most common orientations with other competitors.

\subsection*{PHASE-I}

\begin{table}
\footnotesize
\centering
\begin{tabular}{@{}ccccc@{}}
\toprule
Aggregation & Order Ind. & Base size & precision & recall \\ \midrule
True &  False & 200 & 97.70\% & 63.71\% \\
 &   & 300 & 97.61\% & 70.93\% \\
 &   & 400 & 98.42\% & 76.21\% \\
 &   & 500 & 98.81\% & 78.74\% \\ \cmidrule{2-5}
 &  True & 200 & 98.64\% & 60.12\% \\ 
 &   & 300 & 98.99\% & 69.27\% \\
 &   & 400 & 99.00\% & 74.52\% \\
 &   & 500 & 99.04\% & 77.32\% \\ \midrule
False &  False & 200 & 98.02\% & 60.39\% \\
 &   & 300 & 98.02\% & 68.40\% \\
 &   & 400 & 98.32\% & 73.77\% \\
 &   & 500 & 98.82\% & 76.29\% \\ \cmidrule{2-5}
 &  True & 200 & 98.76\% & 56.45\% \\ 
 &   & 300 & 98.94\% & 66.23\% \\
 &   & 400 & 98.86\% & 71.83\% \\
 &   & 500 & 99.06\% & 75.03\% \\ \bottomrule
\end{tabular}
\caption{Performance based on micro-average of Phase-I.}
\label{tab:p1}
\end{table} 

\begin{table}\centering
\footnotesize
\begin{tabular}{@{}rrrrr@{}} \toprule
Base Size & Aggregation & Order Ind. & TP & FP \\ \midrule
200       & False       & True       & 4.843          & 0.060 \\
          &             & False      & 5.143          & 0.090 \\ \cmidrule{2-5}
          & True        & True       & 5.103          & 0.067 \\
          &             & False      & 5.350          & 0.130 \\ \midrule
500       & False       & True       & 6.373          & 0.057 \\
          &             & False      & 6.493          & 0.093 \\ \cmidrule{2-5}
          & True        & True       & 6.523          & 0.063 \\
          &             & False      & 6.633          & 0.090 \\ \bottomrule
\end{tabular}
\caption{Performance of Phase-I with average number of true positives (TP) and of false positives (FP)/ FPs are reduced to about two thirds by adopting order-independence.}
\label{tab:app:tpfp}
\end{table}

We first report the performance of Robust RpCD for Phase-I. Micro-averaged precision and recall
for undirected dependencies are reported (see \Cref{tab:p1}).
As the size of data increases, more accurate
RCMs are discovered since RCI tests can better catch genuine dependencies.
We observe relatively high precision in general even with a small-sized relational data, which implies that the main problem of the
structure learning is false negatives due to weak dependencies.

\begin{figure}
\centering
\begin{subfigure}{1\columnwidth}
\centering
\includegraphics[width=1\columnwidth]{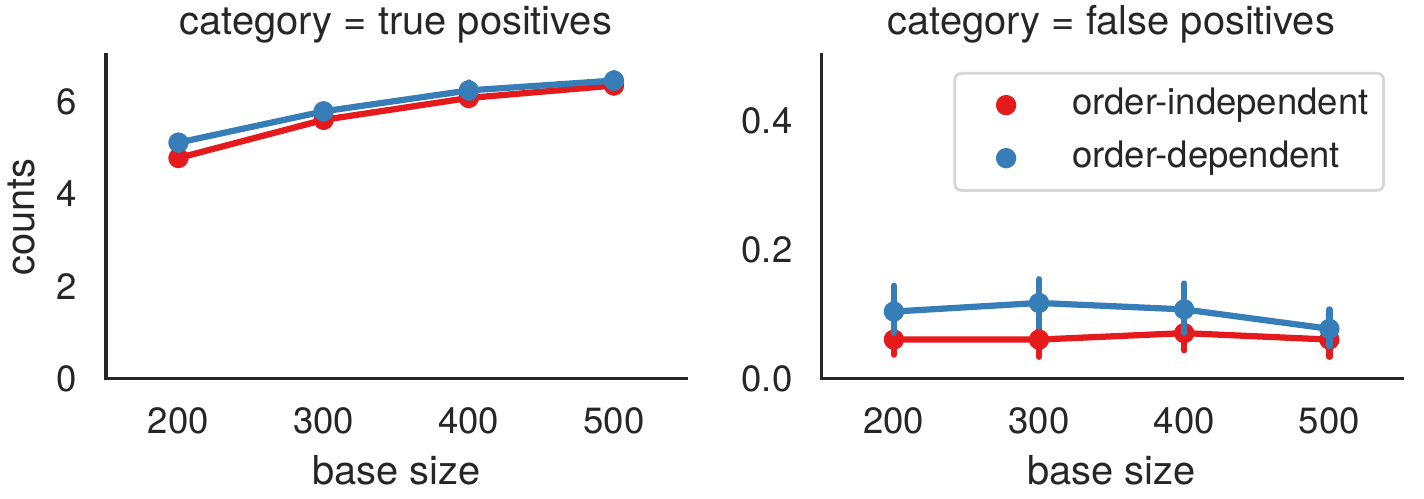}
\caption{Performance on order-independence without aggregation}
\label{fig:perf-order-indep}
\end{subfigure}
\bigskip

\begin{subfigure}{1\columnwidth}
\centering
\includegraphics[width=1\columnwidth]{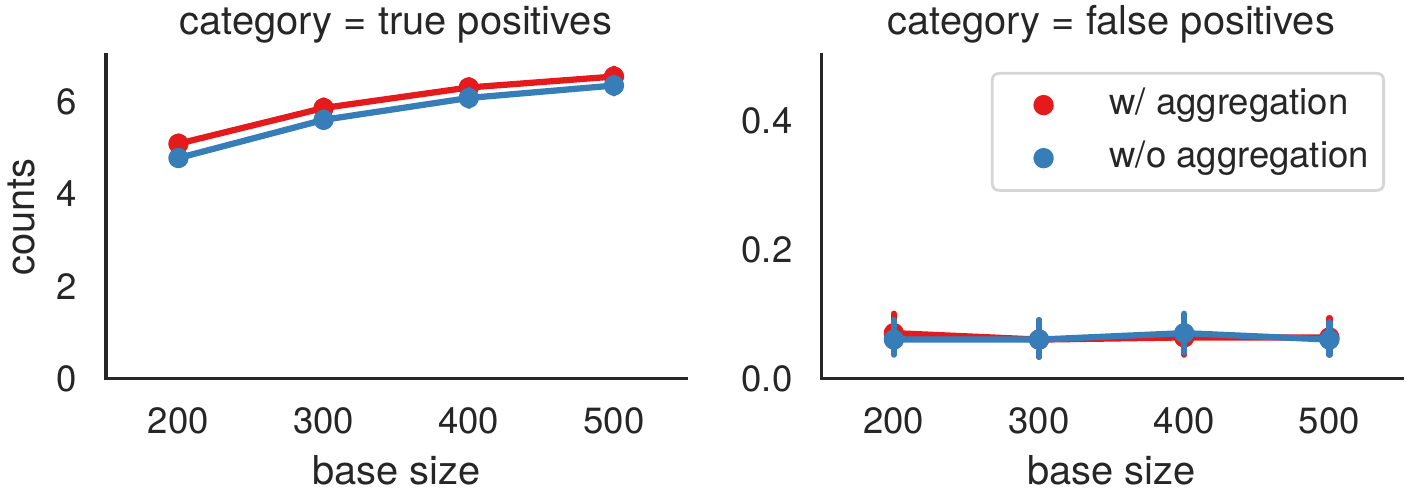}
\caption{Effect of aggregation with order-independence}
\label{fig:rrpcd-effect-aggregation}
\end{subfigure}
\caption{Phase-I}
\end{figure}

\paragraph{Order-independence}
\Cref{fig:perf-order-indep} depicts plots of performance with and without order-independence ---
the average number of true
and false positives without additional
aggregation-based tests. First, order-dependence can yield a higher number
of both true and false positives. We can observe that order-independence
reduces the number of false positives (see \Cref{tab:app:tpfp}).

\begin{figure}
\centering
\includegraphics[width=1\columnwidth]{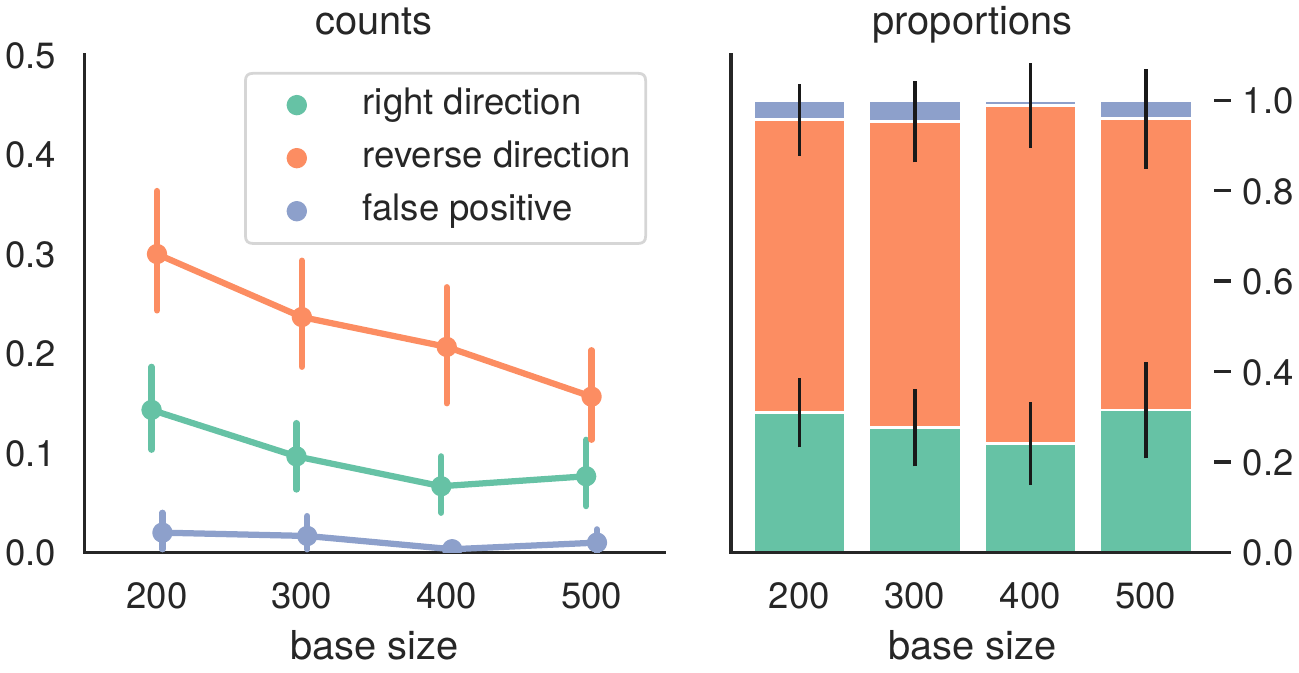}
\caption{\label{fig:aggregation-save}RCI query saved by aggregation-based
tests}
\end{figure}

\paragraph{Aggregation}

In \Cref{fig:rrpcd-effect-aggregation}, aggregation-based CI
tests yield higher true positives without increasing false positives
much. Since the non-aggregated
test and its corresponding aggregated test are correlated, doubling
the test does not significantly increase the false positive rate. 

We explored which types of RCI queries are `saved' by aggregated
tests, i.e., $\left(U\Perp V\mid\mathbf{W}\right)\wedge\left(f\left(U\right)\notperp V\mid\mathbf{W}\right)$
such that $U$ is adjacent to $V$ at the end of Phase I. We report
three cases: i) false positive, $U\not\in adj\left(V;\mathcal{M}\right)$;
ii) right direction, $U\in pa\left(V;\mathcal{M}\right)$; and iii) reverse
direction, $U\in ch\left(V;\mathcal{M}\right)$. We expected that
the aggregation-based test is particularly useful when $U\in ch\left(V;\mathcal{M}\right)$
since $V$ affects each of item attribute in $U$ `individually'.
Then, averaging values might help reducing noises. In \Cref{fig:aggregation-save},
we illustrate the average number of saved dependencies in the three
categories and their proportions. Note that, an adjacency $P.X-\VV{Y}$,
which is also $\tilde{P}.Y-\VV{X}$, can be counted twice. We can first
observe that the total number of saved relational dependencies decreases
as data size increases since the original (i.e., non-aggregation-based)
test will catch weak dependencies better. RCI tests in a reverse direction,
e.g., $U\in ch\left(V;\mathcal{M}\right)$, are mostly saved by aggregation.
The use of aggregation will become more useful as the relationships
in a relational skeleton become more complicated.

\begin{figure}[t]
\scriptsize
\centering
\textsf{Without Detection Mechanisms} \\ [1em]
\includegraphics[width=1\columnwidth]{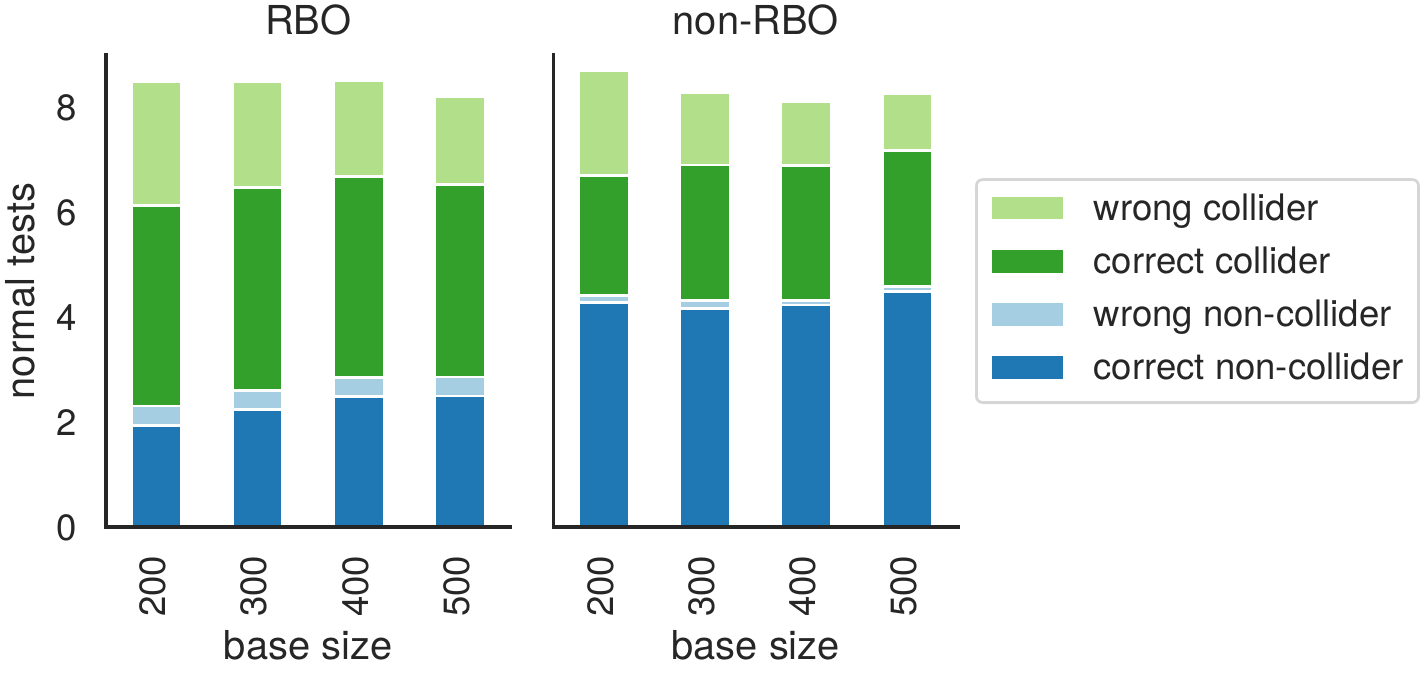}\\ [1em]
\textsf{With Detection Mechanisms} \\ [1em]
\includegraphics[width=1\columnwidth]{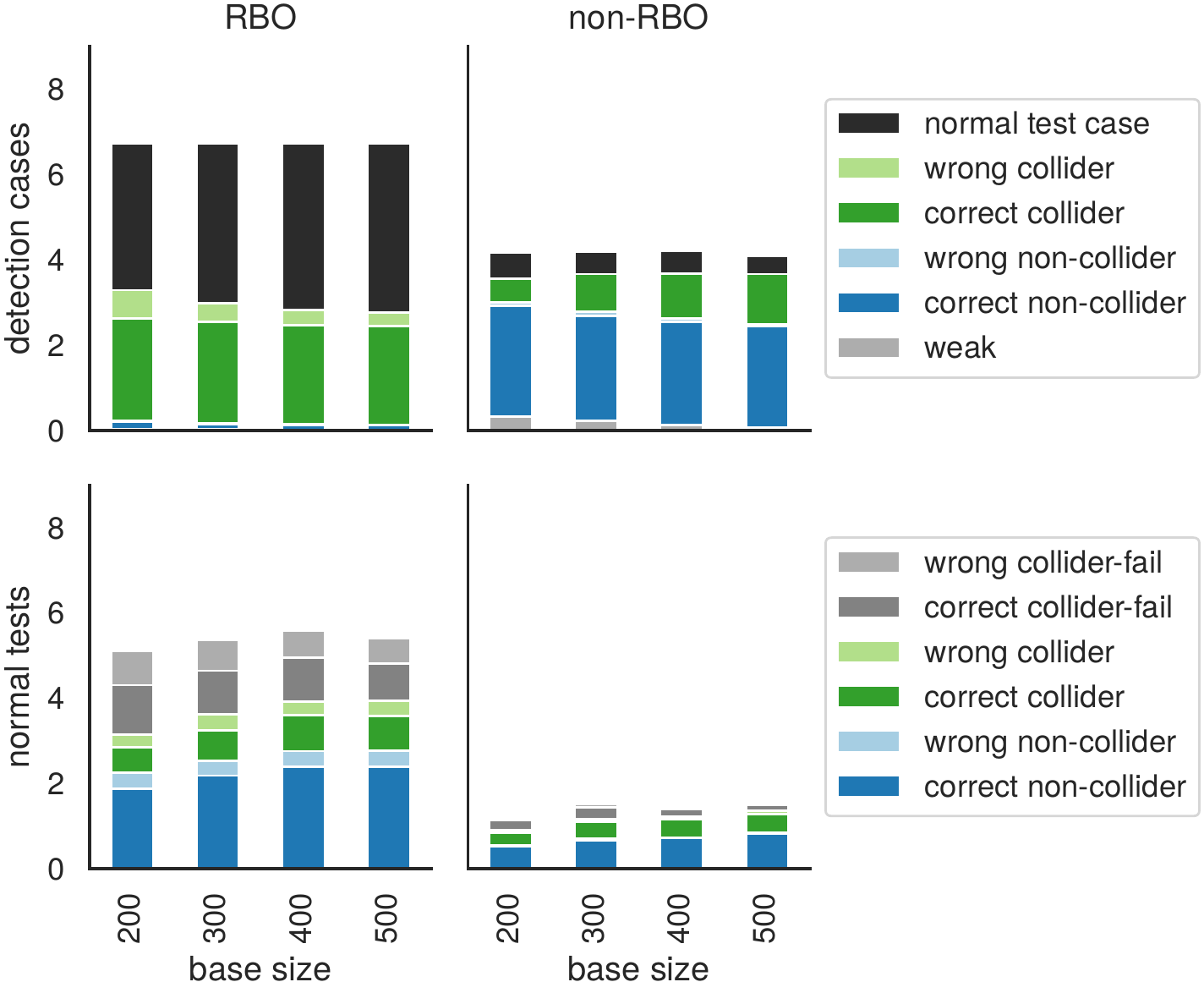}
\caption{Effect of detection mechanism}
\label{fig:rrpcd-detect-p2}
\end{figure}

\subsection*{PHASE-II}

We first overview how each feature affects the performance of orientation
in terms of precision and recall assuming perfect Phase-I, which allows us to judge better how different features work.
More specifically, `correctly directed' relational dependencies
lie in the intersection of oriented relational dependencies through Phase-II and true relational dependencies. Then, precision and recall are the proportion of correctly
directed relational dependencies among directed relational dependencies
through Phase-II, and among directed relational dependencies
in the corresponding CPRCM, respectively.

\begin{table}
\footnotesize
\centering
\begin{tabular}{@{}rrrr@{}}\toprule
Size  & Precision & Recall & F-measure\\ \midrule
200 & 65.8 &  61.0 &  63.3 \\
300 & 74.2 &  67.8 &  70.8 \\
400 & 71.9 &  66.6 &  69.1 \\
500 & 75.3 &  69.7 &  72.4 \\ \bottomrule
\end{tabular}
\caption{Orientation performance of a naive approach with CUT-based RCI tests.}
\label{tab:app:ori:old}
\end{table}

\begin{table}
\footnotesize
\centering
\begin{tabular}{@{}rrrrrr@{}}
\toprule
Agg. & Size & Detection & Prec. & Recall & F\\ \midrule
False       & 200  & False     & 79.0      & 64.5   & 71.0\\
            & 300  &           & 84.7      & 68.7   & 75.8\\
            & 400  &           & 88.1      & 72.8   & 79.7\\
            & 500  &           & 87.8      & 73.6   & 80.1\\ \midrule
False       & 200  & True      & 88.6      & 69.4   & 77.8\\
            & 300  &           & 92.4      & 73.0   & 81.5\\
            & 400  &           & 94.2      & 76.2   & 84.2\\
            & 500  &           & 93.6      & 75.9   & 83.8\\ \midrule
True        & 200  & True      & 88.3      & 70.1   & 78.2\\
            & 300  &           & 91.5      & 73.6   & 81.6\\
            & 400  &           & 93.8      & 76.6   & 84.3\\
            & 500  &           & 93.5      & 75.4   & 83.5\\ \bottomrule
\end{tabular}
\caption{Orientation performance with our proposed approach using RCI tests.}
\label{tab:app:ori:new}
\end{table}

We report micro-average for precision and recall in \Cref{tab:app:ori:old,tab:app:ori:new} for a naive approach (i.e., CUT-based RCI tests with a majority vote rule and a simple sequential strategy to resolve conflicts among orientations.) and our approach (i.e., the proposed RCI tests with the weak dependence detection mechanism, the majority vote rule and a maximal non-conflicting orientations strategy), respectively. The differences in both precision and recall between the two approaches are due to the effectiveness of our proposed RCI tests (as shown in the main text) and the fact that finding a maximally non-conflicting orientations works  as a majority vote rule for final orientations of relational dependencies.


\paragraph{DETECTING CONFLICTS FOR RBO AND NON-RBO}

We investigate how weak dependency detection mechanisms for RBO and non-RBO
work. In \Cref{fig:rrpcd-detect-p2}, we illustrate
the average number of RCI tests which turned out to be colliders or
non-colliders, and whether the RCI test results were right or wrong.

Without detection (the top row), we observe that
there exists a non-negligible amount of wrong collider test results.
This implies that a set of conditionals without blocking $\tilde{P}.Y$
(or $\tilde{Q}.Y$) yields wrong independence. This, again, suggests
how false negatives dominate the performance of the learning algorithm.

With the detection mechanism enabled, the middle row in the figure shows the average orientation results only when an empty set as a separating set is considered. Black bars represent cases where a pair of tests turned out to be dependent, that is, an orientation was not determined. Gray bars (nearly invisible) show cases where both tests returned independence. We can clearly see that the mechanism catches colliders better than without it.

The last row in the figure illustrates orientation results for the undetermined in the previous case (black bars). Note that, since the algorithm seeks for more than one separating set, the lengths of bars in the last row are longer than the lengths of black bars in the middle row.
Collider-fail represents a condition where the detection mechanism rejects a collider since both tests yield independence. More than a half of cases, the mechanism correctly rejected false colliders, yielding a relatively low false collider rate.